\documentclass[a4paper,fleqn]{cas-sc}
\usepackage[numbers,sort&compress]{natbib}
\usepackage{amssymb}
\usepackage{graphicx,nicefrac}
\usepackage{caption}
\usepackage{stmaryrd}
\usepackage{latexsym,amssymb}
\usepackage{amsmath,amsbsy}
\usepackage{amsopn,amstext}
\usepackage{bm}
\usepackage{float}
\usepackage{picins}
\usepackage{subfigure}
\usepackage{graphicx}% Include figure files
\usepackage{color}
\usepackage{longtable}
\usepackage{booktabs}
\usepackage{multirow}
\usepackage{graphicx}
\usepackage{threeparttable}
\usepackage{makecell}
\usepackage{tabularx}
\usepackage{pdfpages}
\newtheorem{definition}{Definition}
\newtheorem{theorem}{Theorem}
\newtheorem{remark}{Remark}
\newenvironment{proof}[1][Proof]{\noindent\textbf{#1.} }{\hfill \rule{0.5em}{0.5em}}
\usepackage{multirow}
\usepackage{algorithm}
\usepackage{algpseudocode}
\usepackage{hyperref}

\definecolor{hyperref-blue}{RGB}{0, 0, 0}
\newcommand{\reffig}[1]{Fig. \ref{#1}}
\newcommand{\refeq}[1]{Eq. \ref{#1}}
\newcommand{\reftable}[1]{Table \ref{#1}}

\def\tsc#1{\csdef{#1}{\textsc{\lowercase{#1}}\xspace}}
\tsc{WGM}
\tsc{QE}
\tsc{EP}
\tsc{PMS}
\tsc{BEC}
\tsc{DE}
\begin{document}
%\textrm{Roman Family}
\let\WriteBookmarks\relax
\def\floatpagepagefraction{1}
\def\textpagefraction{.001}

% Short title
\shorttitle{}    
% Short author
\shortauthors{Jie Hou et ~ al.} 
% Main title of the paper
\title [mode = title]{HNS: An Efficient Hermite Neural Solver for Solving Time-Fractional Partial Differential Equations}  
%
%author
\author[1]{Jie Hou}[type=editor, orcid = 0000-0001-9410-0093]
\ead{houjie@shu.edu.cn}
\credit{Investigation, Methodology, Modeling, Simulation, Validation, Writing - original draft}
%author
\author[2]{Zhiying Ma}[type=editor ]
\ead{mazhiying@shu.edu.cn}
\credit{Modeling, Validation, Review}
%author
\author[2]{Shihui Ying}[type=editor]
\ead{shying@shu.edu.cn}
\credit{Validation, Review}
%author
\author[1]{Ying Li}[type=editor]
\cormark[1]
\ead{yinglotus@shu.edu.cnn}
\cortext[cor1]{Corresponding author}
\credit{Modeling, Validation, Review}
\address[1]{School of Computer Engineering and Science, Shanghai University, Shanghai 200444, PR China}
\address[2]{Department of Mathematics, School of Science, Shanghai University, 200444, PR China}

% Here goes the abstract
\begin{abstract}
Neural network solvers represent an innovative and promising approach for tackling time-fractional partial differential equations by utilizing deep learning techniques.
L1 interpolation approximation serves as the standard method for addressing time-fractional derivatives within neural network solvers. However, we have discovered that neural network solvers based on L1 interpolation approximation are unable to fully exploit the benefits of neural networks, and  the accuracy of these models is constrained to interpolation errors.
In this paper, we present the high-precision Hermite Neural Solver (HNS) for solving time-fractional partial differential equations. 
Specifically, we first construct a high-order explicit approximation scheme for fractional derivatives using Hermite interpolation techniques, and rigorously analyze its approximation accuracy.  
Afterward, taking into account the infinitely differentiable properties of deep neural networks, we integrate the high-order Hermite interpolation explicit approximation scheme with deep neural networks to propose the HNS. The experimental results show that HNS achieves higher accuracy than methods based on the L1 scheme for both forward and inverse problems, as well as in high-dimensional scenarios. This indicates that HNS has significantly improved accuracy and flexibility compared to existing L1-based methods, and has overcome the limitations of explicit finite difference approximation methods that are often constrained to function value interpolation.
 As a result, the HNS is not a simple combination of numerical computing methods and neural networks, but rather achieves a complementary and mutually reinforcing advantages of both approaches. The data and code can be found at \url{https://github.com/hsbhc/HNS}.
\end{abstract} 

\maketitle

%==================================================================================================
\section{Introduction}
%-- reasource allocation is important
Fractional Partial Differential Equations (FPDEs) are an extension of traditional integer-order Partial Differential Equations (PDEs) that stem from the realm of Fractional Calculus, boasting rich research backgrounds in mathematics and physics \cite{podlubny1999introduction, kilbas2006theory}.
Compared to integer-order PDEs, fractional derivatives can more accurately represent phenomena with memory and hereditary attributes.
As a result, FPDEs can effectively capture memory and non-local effects in various natural phenomena.  This makes them highly valuable in fields such as fluid dynamics \cite{li2021fractional}, turbulence \cite{song2016fractional}, geology \cite{beltempo2018fractional}, elasticity \cite{mainardi2022fractional}, biology \cite{ionescu2017role}, financial theory \cite{tarasov2019history,chen2020combined}, and engineering \cite{sun2018new}.

In recent decades, researchers have made significant progress in the numerical solution of fractional-order differential equations, developing numerous numerical techniques. However, due to the inherent nonlinearity, high dimensionality, and complex boundary conditions of FPDEs, the computational complexity of traditional numerical methods is often significant\cite{sousa2009finite,sweilam2007numerical,lin2007finite,bhrawy2014new,kazem2013fractional,jafari2011application,langlands2005accuracy}. Furthermore, traditional numerical methods may encounter difficulties in effectively addressing the memory effects and long-range dependencies inherent in FPDEs, which can result in diminished accuracy in their computational results. When tackling high-dimensional problems, numerical methods typically involve extensive computational demands\cite{gu2020meshless}.

With the growing development of deep neural networks, it is recognized that neural networks have powerful approximation and nonlinear modeling capabilities \cite{lecun2015deep,bishop1994neural}, which makes neural networks show great potential in solving scientific computing problems \cite{dl_PDE_karniadakis2021physics,dl_PDE_lu2021deepxde,dl_PDE_raissi2020hidden}.  Recently, the Physics-Informed Neural Networks (PINNs) model has achieved notable success in solving Partial Differential Equations (PDEs) \cite{PINN_raissi2019physics} and has been applied to FPDEs. 
Pang et al.\cite{pang2019fpinns} extended PINNs to Fractional PINNs (fPINNs), exhibiting accuracy and efficacy in solving forward and inverse problems while systematically exploring convergence properties. 
Ye et al.\cite{ye2022deep} conducted a study on time-fractional diffusion equations with conformable derivative using Physics-Informed Neural Networks (PINNs). For more instructions on neural network methods for solving fractional-order differential equations, see \cite{pakdaman2017solving,qu2022neural,biswas2023study,wang2023physics,ye2023slenn,hajimohammadi2021fractional,yan2023laplace}. The primary advantage of neural network methods lies in their ability to solve high-dimensional problems, handle complex geometries, and address inverse problems effectively.

The main challenge in solving fractional partial differential equations (FPDEs) lies in the handling of fractional derivatives. Similar to many neural network approaches for solving PDEs, it is common to combine numerical methods to efficiently solve problems\cite{chiu2022can,sharma2022accelerated}. Therefore, when solving FPDEs, we also use traditional approximation methods to approximate fractional-order derivatives. As a result, the accuracy of neural network models in solving FPDEs is closely related to the choice of approximation methods for fractional derivatives. Therefore, to ensure the accuracy and stability of the model, it is crucial to select appropriate numerical approximation methods.
Most current methods employ finite difference approximation methods to discretize the time-fractional derivatives, with the accuracy of the model being limited by the scheme approximation errors. For example, when discretizing time fractional derivatives using the finite difference L1 scheme directly \cite{sun2006fully,jin2016analysis}, the accuracy of the L1 scheme determines the prediction accuracy of the model.

In fact, directly combining finite difference approximation methods for fractional derivatives with deep neural networks is not an ideal choice. Such a simplistic combination does not fully leverage the advantages of neural networks, and the accuracy of the model is limited to the accuracy of numerical approximation methods.
Therefore, when designing integration strategies, we should fully leverage the potential of neural networks to enhance the overall model accuracy.
For instance, we commonly use the L1 scheme to approximate Caputo fractional-order derivatives. This approximation method relies on function values for linear interpolation, which results in significant interpolation errors. This is a feature of explicit difference approximation methods, even with higher-order difference schemes\cite{cao2015high,ying2017high}$-$This is because they are explicit methods that construct interpolation polynomials exclusively through function values, without taking into account other aspects of the function, such as the derivative value of the nodes. Is it possible to break through this limitation by the neural networks? The answer is affirmative.

In this paper, we utilize the {\bf Hermite interpolation method}\cite{suli2003introduction} to construct a high-order explicit approximation scheme for the Caputo fractional derivative and design an efficient Hermite Neural Solver for solving time fractional partial differential equations. This method fully utilizes {\bf the infinitely differentiable properties} of deep neural networks, breaks through the limitation of constructing interpolation polynomials only through function values, and improves numerical accuracy. Moreover, this method can easily extend to other fractional partial differential equations, even integral equations. Although there are cubic spline collocation methods available for solving FPDEs outside of explicit finite difference methods\cite{zahra2012use,yang2013cubic,pitolli2022approximation}, these methods utilize cubic spline interpolation techniques similar to the Hermite interpolation method. However, such methods introduce additional degrees of freedom, making them implicit iterative methods. In contrast, this paper leverages neural networks as a tool, allowing us to explicitly execute {\bf Hermite Interpolation Explicit Approximation for Caputo Fractional Derivatives}. To the best of our knowledge, this is the first work that combines deep neural networks with the Hermite interpolation method and designs a high-order Hermite Neural Solver for solving fractional partial differential equations. The main contributions of this paper are summarized as follows:
\begin{itemize}
\item A high-order explicit approximation scheme for the Caputo fractional derivative is constructed based on Hermite interpolation technique, and its approximation error is analyzed.
\item An efficient Hermite Neural Solver (HNS) is designed for solving time fractional partial differential equations.
\item The impact of Hermite interpolation order on the accuracy of the HNS is analyzed. The results show that the HNS using third-order Hermite interpolation achieves the highest performance.
\item The performance of HNS has been thoroughly verified across various problem scenarios, including challenging high-dimensional problems and inverse problems.
\end{itemize}

\section{Preliminaries}
\subsection{Time-Fractional Partial Differential Equation}
Without loss of generality, we formulate any Time-Fractional PDEs as:
\begin{equation}\label{eq:e1}
	\frac{\partial^{\alpha} u(\boldsymbol{x},t)}{\partial t^{\alpha}}+ \mathcal{N}_{\boldsymbol{x}}[u] = 0,
\end{equation}
where $u(\boldsymbol{x},t)$ is the function which needs to be solved, $\alpha$ is the order of the time derivative, and $\mathcal{N}_{\boldsymbol{x}}[\cdot]$ is a differential operator that can be linear or nonlinear. Fractional derivatives are an extension of calculus that allow for derivative orders to be real numbers rather than limited to integers. This extension finds widespread application in the fields of mathematics, physics, and engineering, particularly when describing complex systems characterized by memory and non-local properties. Commonly used definitions of fractional derivatives include Riemann-Liouville fractional derivative and Caputo fractional derivative \cite{miller1993introduction}. In this paper, we employ the definition of Caputo fractional derivatives. 
\begin{definition}
\label{definition1}
	Suppose the function $u(t)$ is defined on the interval $[a, b]$, $\alpha\geq 0$, and $n$ is the smallest integer greater than or equal to $\alpha$. The Caputo fractional derivative is defined as:
\begin{equation}
	{ }_{a}^{C} D_{t}^{\alpha} u(t)=\left\{
	\begin{aligned}
&\frac{1}{\Gamma(n-\alpha)} \int_{a}^{t}(t-\tau)^{n-\alpha-1} u^{(n)}(\tau) d \tau,& \alpha \in (n-1,n), \\
&u^{(n)}(t), &\alpha = n \in N, \ \ \ \
\end{aligned}
\right.
\end{equation}
where $\Gamma(\cdot)$ denotes the Gamma function. 
\end{definition}
If $t>0$ and $\alpha\in(0,1)$, the Caputo fractional derivative can be written as:
\begin{equation}
	{ }_{0}^{C} D_{t}^{\alpha} u(t)=
\frac{1}{\Gamma(1-\alpha)} \int_{0}^{t} \frac{u^{\prime}(\tau)}{(t-\tau)^{\alpha}} d \tau.
\end{equation}
This paper focuses on this case. To solve the problem \refeq{eq:e1}, numerical methods are needed to discretize the fractional derivative. For the Caputo fractional derivative, finite difference approximation methods, such as the L1 scheme, are commonly used for temporal discretization. Spatial discretization can be achieved using methods such as finite differences, spectral methods, finite elements, and so on.
\paragraph{L1 Scheme}
Let $t\in(0,T]$, $\alpha\in(0,1)$ and $N$ be a positive integer. Denote $\Delta t=\frac{T}{N}$, $t_k=k\Delta t$ for $0\leq k\leq N$, and $a_l=(l+1)^{1-\alpha}-l^{1-\alpha}$ for $l\geq 0$. Linear interpolation of $u(t)$ over the interval $[t_{k-1}, t_k]$ yields:
\begin{equation}
	u(t)\approx L_{k}(t) = \frac{t_{k}-t}{\Delta t}u(t_{k-1})+\frac{t-t_{k-1}}{\Delta t}u(t_{k}). 
\end{equation}
Then, we can obtain the following L1 approximation scheme for computing ${ }_{0}^{C} D_{t}^{\alpha} u(t)\mid_{t=t_{n}}$:
\begin{equation}
\label{eq:e5}
	D_{t}^{\alpha} u(t_{n})=\frac{\Delta t^{-\alpha}}{\Gamma(2-\alpha)}\left[a_{0}u(t_{n}) - \sum_{k=1}^{n-1}(a_{n-k-1} - a_{n-k})u(t_{k})-a_{n-1}u(t_{0})\right].
\end{equation}
The approximation error of the fractional derivative can be expressed as:
\begin{equation}
	R(u(t_{n})) = { }_{0}^{C} D_{t}^{\alpha} u(t)\mid_{t=t_{n}} - D_{t}^{\alpha} u(t_{n}).
\end{equation}
L1 scheme has a numerical accuracy of $2-\alpha$ order. The approximation error satisfies:
\begin{equation}
	\mid R(u(t_{n})) \mid \leq \frac{1}{2\Gamma(1-\alpha)}\left[\frac{1}{4} + \frac{\alpha}{(1-\alpha)(2-\alpha)}\right]\underset{t_{0}\leq t \leq t_{n}}{\operatorname{max}}
	\left|u^{\prime\prime}(t)\right| \Delta t^{2-\alpha}.
\end{equation}
\begin{remark}
	The accuracy of explicit finite difference approximation methods is limited by the interpolation error. A similar approximation method is the $L1-2$ interpolation approximation. This method employs linear interpolation of $u(t)$ on the interval $[t_{0}, t_{1}]$ and uses a three-point quadratic interpolation polynomial in subsequent intervals. The $L1-2$ approximation formula, as the number of computational nodes increases, transitions from being of order $2-\alpha$ to order $3-\alpha$. So, can we employ alternative higher-order interpolation methods, such as Hermite interpolation, to achieve a high-precision approximation in conjunction with finite differences? 
\end{remark}

\subsection{Hermite Interpolation}
Hermite interpolation is an interpolation method that requires the interpolating polynomial to pass through the given nodes and match the function values as well as the values of the first several derivatives of the original function at each node \cite{suli2003introduction}.
\begin{definition}
\label{definition 2}
	Let $f(t)$ be a function defined on $[a,b]$, and let $t_0,t_1,\ldots,t_n$ be $n+1$ distinct points in $[a,b]$, where the function values $f(t_i)$ and the values of the first $k$ derivatives $f'(t_i),\ldots,f^{(k)}(t_i)$ are all known. The Hermite interpolation polynomial $H(t)$ satisfies the following conditions:
\begin{itemize}
	\item $H(t)$ passes through each given point $t_i$ and takes on the corresponding function value, i.e., $H(t_i) = f(t_i)$.
	\item $H(t)$ has the same values as $f(t)$ for the first $k$ derivatives at each given point $t_i$, i.e., $H^{(j)}(t_i) = f^{(j)}(t_i)$, for $1 \leq j \leq k$.
\end{itemize}
\end{definition} 
\begin{theorem}
The Hermite polynomial of degree $p$, denoted as $H_{p}(t)$, and the original function $f(t)$ are under consideration. When evaluating at a point $t$ within the interval $[t_{0}, t_{n}]$, the error function can be expressed as follows:
\begin{equation}
	f(t)-H_{p}(t)=\frac{f^{(p+1)}(\xi)}{(p+1) !} \prod_{i}\left(t-t_{i}\right)^{\frac{p+1}{2}},
\end{equation}
where  $\xi$ is an unknown point in the range $ \left(t_{0}, t_{n}\right)$, $p+1$  is the total number of data points.
\end{theorem}

\paragraph{Two-Point  Hermite Interpolation} Two-Point Hermite interpolation is a commonly employed method in numerical analysis and interpolation problems. 
Given an interval $[a,b]$ and the values of a function $f(a)$, $f'(a)$, $f''(a)$ and $f(b)$, $f'(b)$, $f''(b)$, we can construct a linear interpolation function $H_{1}(t) = h^{0}_{1}(t)f(a) + h^{1}_{1}(t)f(b)$ using only the function values $f(a)$ and $f(b)$. The subscripts on $H_{p}(t)$ and $h_{p}(t)$ denote that they are polynomials of degree $p$. According to {\bf Definition \ref{definition 2}}, $h^{0}_{1}(t)$ and $h^{1}_{1}(t)$ must satisfy:
\begin{equation}
	\begin{aligned}
		h^{0}_{1}(a)=1,h^{0}_{1}(b)=0,\\
		h^{1}_{1}(a)=0,h^{1}_{1}(b)=1.
	\end{aligned}
\end{equation}
Thus $h^{0}_{1}(t) = \frac{t-b}{a-b}$ and $h^{1}_{1}(t) = \frac{a-t}{a-b}$, which gives:
\begin{equation}\label{eq:e10}
	H_{1}(t) = \frac{t-b}{a-b}f(a) + \frac{a-t}{a-b}f(b).
\end{equation}
This is equivalent to the linear interpolation used in the {\bf L1 scheme} for the Caputo fractional derivative. 
The error of this interpolation approximation is:
\begin{equation}
	f(t)-H_{1}(t)=\frac{f^{\prime\prime}(\xi)}{2!}\left(t-a\right)\left(t-b\right), \ \ \xi \in (a,b).
\end{equation}
Similarly, it is possible to construct a function by using its values and first-order derivative values,
\begin{gather}\label{eq:e11}
	H_{3}(t)= h^{0}_{3}(t)f(a) + h^{1}_{3}(t)f(b)+h^{2}_{3}(t)f^{\prime}(a) + h^{3}_{3}(t)f^{\prime}(b)\\
	f(t)-H_{3}(t)=\frac{f^{(4)}(\xi)}{4!}\left(t-a\right)^{2}\left(t-b\right)^{2}, \ \ \xi \in (a,b).
\end{gather}
Incorporating constraints on the second-order derivative values can obtain:
\begin{gather}\label{eq:e12}
	H_{5}(t ) = h^{0}_{5}(t)f(a) + h^{1}_{5}(t)f(b)+h^{2}_{5}(t)f^{\prime}(a) + h^{3}_{5}(t)f^{\prime}(b)+h^{4}_{5}(t)f^{\prime\prime}(a) + h^{5}_{5}(t)f^{\prime\prime}(b)\\
		f(t)-H_{5}(t)=\frac{f^{(6)}(\xi)}{6!}\left(t-a\right)^{3}\left(t-b\right)^{3}, \ \ \xi \in (a,b). 
\end{gather}
While higher-degree Hermite interpolation polynomials theoretically yield smaller errors, they are often vulnerable to the impact of rounding errors, which can lead to numerical instability. Additionally, they may exhibit Runge's phenomenon, causing oscillations within the interval between interpolation nodes. As a result, in practical scenarios, piecewise cubic Hermite interpolation polynomials are commonly favored.
\begin{remark}
	Now, let us attempt to address the question posed in {\bf Remark 1}. Can we replace the interpolation step in finite difference approximation methods with Hermite interpolation to achieve higher precision in fractional derivative approximations? In traditional numerical methods, even without considering issues of regularity in the solution, direct replacement is not feasible because both the function values and derivative values of the solution are unknown. Using Hermite interpolation would require introducing additional degrees of freedom and solving the problem through implicit iterations. However, it is worth noting that in neural network approaches, it appears that we can directly substitute the interpolation step to achieve this goal. Since the implicit iteration steps are inherently embedded within the neural network optimization process, we can directly execute high-order interpolation approximation methods.
\end{remark}

\section{Methodology}
\subsection{A Review of Neural Networks for Solving PDEs and FPDEs}
This section reviews the applications of neural networks in solving partial differential equations (PDEs) and fractional partial differential equations (FPDEs). We will focus on two main methods: Physics-Informed Neural Networks (PINNs) and Fractional Physics-Informed Neural Networks (fPINNs).
\paragraph{Physics-informed Neural Networks (PINNs)} PINNs are a type of models that incorporates prior knowledge of physics into the neural network training process \cite{PINN_raissi2019physics}. It can infer a continuous solution function $u(\boldsymbol{x}, t)$ based on physical equations.
Consider a general nonlinear partial differential equation, which is given by:
\begin{equation}
	\begin{array}{l}
u_{t}+\mathcal{N}_{\boldsymbol{x}}[u]=0,\ \ \quad \boldsymbol{x} \in \Omega, t \in[0, T], \\
u(\boldsymbol{x}, 0)=I(\boldsymbol{x}), \ \ \ \quad \boldsymbol{x} \in \Omega, \\
u(\boldsymbol{x}, t)=B(\boldsymbol{x}, t), \quad\boldsymbol{x} \in \partial \Omega, t \in[0, T], 
\end{array}
\end{equation}
where $\mathcal{N}_{\boldsymbol{x}}[\cdot]$ is a nonlinear differential operator, $T$ is the time range. Given the initial condition $u(\boldsymbol{x}, 0)=I(\boldsymbol{x})$ and the boundary condition $u(\boldsymbol{x}, t)=B(\boldsymbol{x}, t),\ \boldsymbol{x} \in \partial \Omega$, our aim is to find the solution function $u(\boldsymbol{x}, t)$ under these known conditions. Firstly, we construct a trial solution:
 \begin{equation}
 	\tilde{u}(\boldsymbol{x}, t) = t f(\boldsymbol{x}, t;\theta) + I(\boldsymbol{x}),
 \end{equation}
 where $f(\boldsymbol{x}, t;\theta)$  is a fully connected neural network, $\theta$ denotes the set of parameters for the neural network. $\tilde{u}(\boldsymbol{x}, t)$ naturally satisfies the initial conditions. Secondly, we define the residual of the PDE as:
\begin{equation}
	r(\boldsymbol{x}, t):=\tilde{u}_{t}(\boldsymbol{x}, t)+\mathcal{N}_{\boldsymbol{x}}\left[\tilde{u}(\boldsymbol{x}, t)\right],
\end{equation}
where the partial derivatives of $\tilde{u}(\boldsymbol{x}, t)$ can be effortlessly obtained utilizing automatic differentiation. The loss function of the PINNs model can be expressed as:
\begin{equation}
	\mathcal{L}(\theta)= \frac{1}{N_{b}} \sum_{j=1}^{N_{b}}\left[\tilde{u}(\boldsymbol{x}_{b}^{j}, t_{b}^{j})-B(\boldsymbol{x}_{b}^{j}, t_{b}^{j})\right]^{2}
	 +\frac{1}{N_{r}} \sum_{j=1}^{N_{r}}\left[r(\boldsymbol{x}_{r}^{j}, t_{r}^{j})\right]^{2},
\end{equation}
where $\{(\boldsymbol{x}_{b}^{j}, t_{b}^{j}), B(\boldsymbol{x}_{b}^{j}, t_{b}^{j})\}_{j=1}^{N_{b}}$ represents the boundary point data, and $\{\boldsymbol{x}_{r}^{j}, t_{r}^{j}\}_{j=1}^{N_{r}}$ refers to the internal collocation points, $N_{b}, N_{r}$ are the respective number of data points. The internal collocation points are uniformly sampled coordinate points from the domain $\Omega$ to compute the residual loss, which enforces $\tilde{u}(\boldsymbol{x}, t)$ to satisfy the governing equation. By continuously adjusting the neural network parameters to minimize the loss function and bringing $\mathcal{L}(\theta)$ as close to zero as possible, we can consider $\tilde{u}(\boldsymbol{x}, t)$ as the approximate solution function when the loss decreases to a minimal value.

\paragraph{Fractional PINNs (fPINNs)} 
fPINNs are an extension of PINNs, used for solving FPDEs \cite{pang2019fpinns}. By replacing $u_t$ with ${ }_{0}^{C} D_{t}^{\alpha} u$, where the time derivative is no longer an integer but a fractional order $\alpha$, the problem becomes a time-fractional partial differential equation. The residual of the FPDE can be expressed as:
\begin{equation}
	r(\boldsymbol{x}, t):={ }_{0}^{C} D_{t}^{\alpha}\tilde{u}(\boldsymbol{x}, t) +\mathcal{N}_{\boldsymbol{x}}\left[\tilde{u}(\boldsymbol{x}, t)\right].
\end{equation}
Unlike integer-order derivatives, fractional derivatives cannot be obtained directly through automatic differentiation. Therefore, in fPINNs, the time-fractional derivative is approximated using the finite difference L1 scheme, as shown in \refeq{eq:e5}.
 Given the spatial location $\boldsymbol{x}$, ${ }_{0}^{C} D_{t}^{\alpha}\tilde{u}(\boldsymbol{x}, t) $ evaluated at time $t$ depends on all the values of $\tilde{u}(\boldsymbol{x}, t)$ evaluated at all the
previous time steps $0, \Delta t, 2\Delta t, \cdots , t$.

\subsection{Time-Fractional Hermite Neural Solver (HNS)}
A simple way to solve time-fractional PDEs is to combine finite difference approximation methods for fractional derivatives with PINNs. However, explicit finite difference approximation methods can only construct interpolation functions based on function values, resulting in low performance. In this paper, we introduce a novel neural network method for solving time-fractional PDEs, utilizing Hermite interpolation.

\begin{figure}[htbp]
  \centering
  \includegraphics[width=1\textwidth]{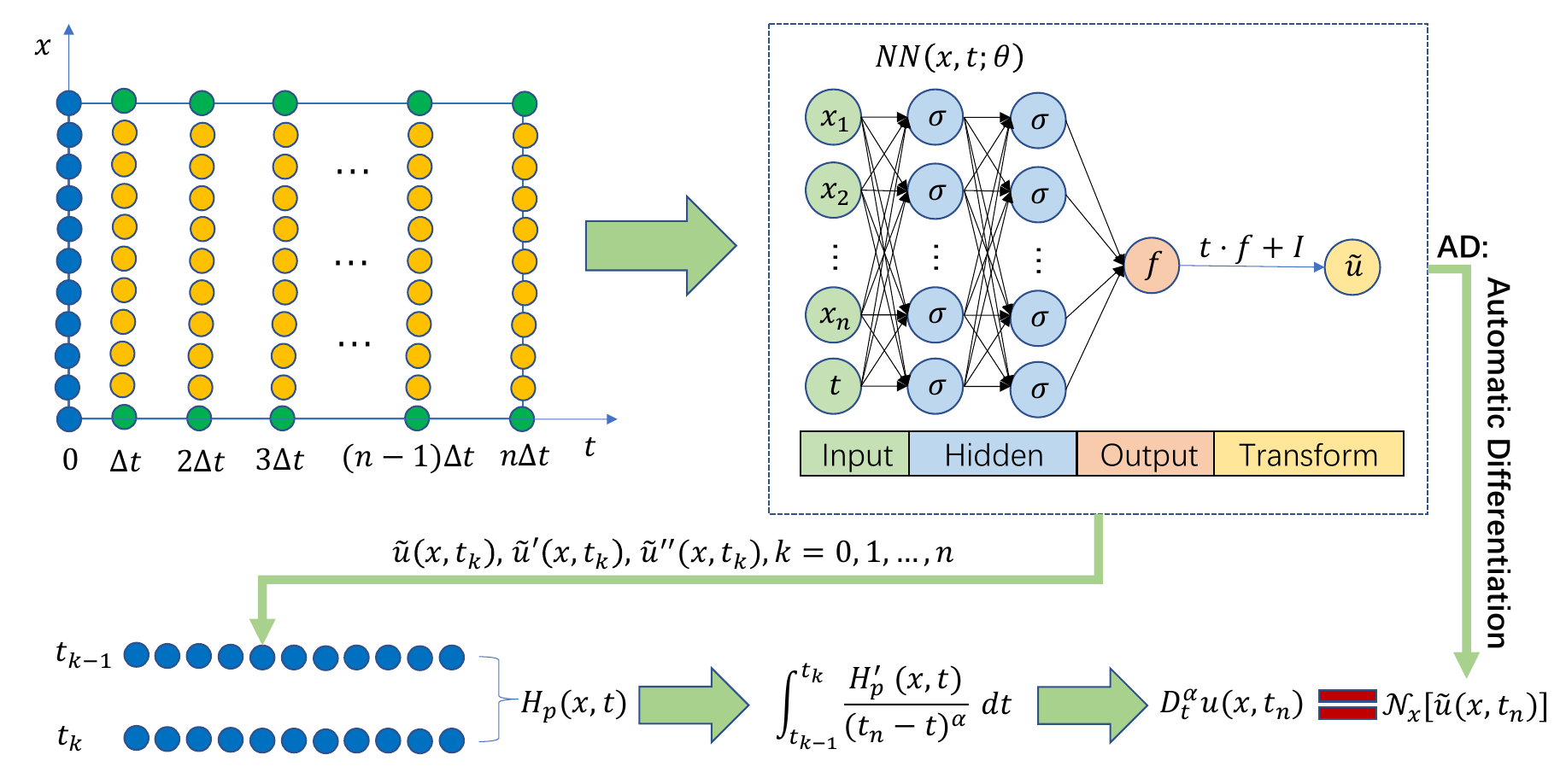}
  \caption{HNS for 1D problem: Blue dots are known points, green dots are boundaries, and yellow dots are sampling points. After inputting them into the neural network, they become known points (blue) for calculating Caputo fractional-order derivatives using Hermite interpolation approximation.}
  \label{TF-HNS}
\end{figure}

\begin{figure}[htbp]
  \centering
  \includegraphics[width=1\textwidth]{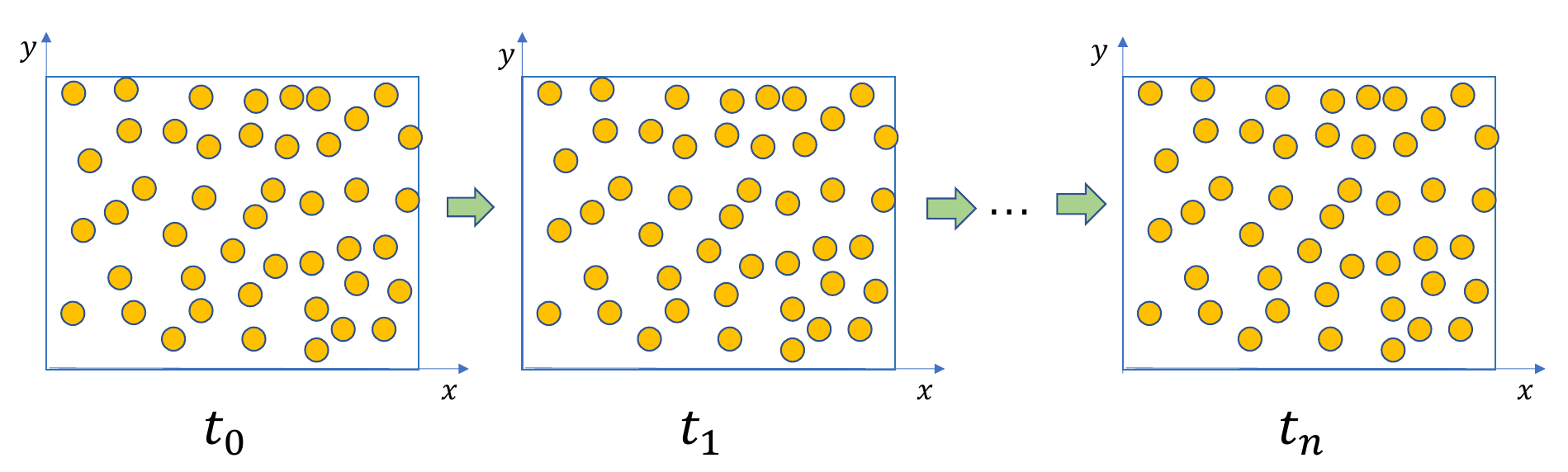}
  \caption{HNS: the spatial coordinates can be sampled uniformly or equidistantly.}
  \label{sampled}
\end{figure}

Firstly, we construct a high-order explicit approximation scheme for the fractional derivative based on Hermite interpolation. We then use the high-order explicit approximation scheme to compute residuals for optimizing network parameters. This method takes advantage of the infinitely differentiable properties of neural networks and extends the finite difference approximation methods. This fully leverages the power of neural networks. In addition, this method does not require implicit iterations; instead, it utilizes neural networks to perform explicit, higher-order interpolation for approximating fractional derivatives. As a result, it offers computational efficiency similar to finite difference methods while achieving higher accuracy.
Specifically, for the Caputo time-fractional PDEs given by:
\begin{equation}
{ }_{0}^{C} D_{t}^{\alpha} u(\boldsymbol{x}, t)+\mathcal{N}_{\boldsymbol{x}}[u]=0,\ \ \quad \boldsymbol{x} \in \Omega, t \in[0, T],
\end{equation}
where $\alpha$ is the order of the time derivative and we focus on the case where $\alpha \in (0,1)$. The solution function $u(\boldsymbol{x}, t)$ must satisfy the initial condition $u(\boldsymbol{x}, 0)=I(\boldsymbol{x})$ and the boundary condition $u(\boldsymbol{x}, t)=B(\boldsymbol{x}, t),\ \boldsymbol{x} \in \partial \Omega$. Firstly, we construct a trial solution for this problem based on the initial condition:
 \begin{equation}
 	\tilde{u}(\boldsymbol{x}, t) = t f(\boldsymbol{x}, t;\theta) + I(\boldsymbol{x}),
 \end{equation}
 where $\tilde{u}(\boldsymbol{x}, t)$ satisfies the initial condition without any additional constraints. Subsequently, we define the residual and the loss function as:
 \begin{gather}
 	r(\boldsymbol{x}, t):={ }_{0}^{C} D_{t}^{\alpha}\tilde{u}(\boldsymbol{x}, t)+\mathcal{N}_{\boldsymbol{x}}\left[\tilde{u}(\boldsymbol{x}, t)\right],\\
 	\mathcal{L}(\theta)= \frac{1}{N_{b}} \sum_{j=1}^{N_{b}}\left[\tilde{u}(\boldsymbol{x}_{b}^{j}, t_{b}^{j})-B(\boldsymbol{x}_{b}^{j}, t_{b}^{j})\right]^{2}
	 +\frac{1}{N_{r}} \sum_{j=1}^{N_{r}}\left[r(\boldsymbol{x}_{r}^{j}, t_{r}^{j})\right]^{2},
 \end{gather}
where $\mathcal{N}_{\boldsymbol{x}}\left[\tilde{u}(\boldsymbol{x}_{r}^{j}, t_{r}^{j})\right]$ can be obtained through automatic differentiation, while ${ }_{0}^{C} D_{t}^{\alpha}\tilde{u}(\boldsymbol{x}_{r}^{j}, t_{r}^{j})$ will be approximated using {\bf Hermite Interpolation Explicit Approximation}. Due to the non-local nature of fractional derivatives, as defined in {\bf Definition \ref{definition1}}, evaluating $r(\boldsymbol{x}, t)$ using random points in the computational domain requires the introduction of auxiliary nodes from previous time steps, which results in lower computational efficiency. To improve computational efficiency, we use equidistant nodes on the time axis to evaluate $r(\boldsymbol{x}, t)$, i.e., $\left\{\{\boldsymbol{x}_{r}^{j}, 0\}_{j=1}^{M_x}\cup\{\boldsymbol{x}_{r}^{j}, \Delta t\}_{j=1}^{M_x} \cup  \cdots \cup \{\boldsymbol{x}_{r}^{j}, T\}_{j=1}^{M_x}\right\}$, where $M_x$ is the number of spatial coordinates sampled in the spatial domain $\Omega$, and the spatial coordinates can be sampled uniformly or equidistantly, see \reffig{sampled}. In addition, in this article, we utilize the L-BFGS optimizer to optimize the network parameters. The model architecture is shown in \reffig{TF-HNS}.

\paragraph{Hermite Interpolation Explicit Approximation for Caputo Fractional Derivatives}
In explicit finite difference approximation methods, we can only construct interpolation functions based on function values, which greatly limits the accuracy of numerical methods. In this paper, we leverage the infinitely differentiable properties of neural networks and use Hermite interpolation methods to construct high-order interpolation explicit approximation methods based on both function and derivative values. This paper primarily investigates time-fractional partial differential equations, So let us focus on the temporal dimension. For clarity, we will use the example of the fractional ordinary differential equation to illustrate the Hermite Interpolation Explicit Approximation for Caputo Fractional Derivatives. To utilize Hermite interpolation, it is necessary to assume the regularity of the solution, by assuming that $u(t) \in C^{p+1}[0,T]$.
\paragraph{Hermite Interpolation Explicit Approximation} Let $t \in (0,T]$ and $\alpha \in (0,1)$ be given. Take a positive integer $N$. Let $\Delta t = \frac{T}{N}$ and $t_{k} = k\Delta t$, $0\leq k\leq N$. Then, we have:
\begin{equation}
	{ }_{0}^{C} D_{t}^{\alpha} u(t)\mid_{t=t_{n}}=
\frac{1}{\Gamma(1-\alpha)} \int_{0}^{t_{n}} \frac{u^{\prime}(t)}{(t_{n}-t)^{\alpha}} dt =  \frac{1}{\Gamma(1-\alpha)} \sum_{k=1}^{n}\int_{t_{k-1}}^{t_{k}} \frac{u^{\prime}(t)}{(t_{n}-t)^{\alpha}} dt.
\end{equation}
On the interval $[t_{k-1},t_{k}]$, we perform Hermite interpolation on $u(t)$ using formulas \refeq{eq:e10}, \refeq{eq:e11}, \refeq{eq:e12} to obtain:
\begin{equation}
	u(t)\approx H_{p}(t), 
\end{equation}
Thus, we obtain the Hermite approximation scheme for computing ${ }_{0}^{C} D_{t}^{\alpha} u(t)\mid_{t=t_{n}}$:
\begin{equation}
	D_{t}^{\alpha} u(t_{n})=\frac{1}{\Gamma(1-\alpha)} \sum_{k=1}^{n}\int_{t_{k-1}}^{t_{k}} \frac{H^{\prime}_{p}(t)}{(t_{n}-t)^{\alpha}} dt,
\end{equation}
where $H_{p}(t)$ is a polynomial of degree $p$ and $H^{\prime}_{p}(t)$ is a polynomial of degree $p-1$. The integration operation can be obtained using {\bf symbolic computation} to obtain an exact expression in advance.

\begin{theorem}
	Suppose $u(t) \in C^{p+1}[0,T]$, then the approximation error is given by:
\begin{equation}
\begin{aligned}
	&\mid R(u(t_{n})) \mid \\
	&\leq \frac{1}{(p+1) !\Gamma(1-\alpha)} \left[ \frac{1}{2^{p+1}} + \alpha \frac{\Gamma(\frac{3+p}{2}) \Gamma(\frac{1-2\alpha+p}{2})}{\Gamma(2-\alpha+p)} \right]\underset{t_{0}\leq \xi \leq t_{n}}{\operatorname{max}} \left|u^{(p+1)}(\xi)\right| \Delta t^{p+1-\alpha}
\end{aligned}
\end{equation}
\end{theorem}
The proof can be found in Appendix \ref{Proof}.

\begin{remark}
	The Hermite interpolation Explicit approximation for Caputo fractional derivatives theoretically has a numerical accuracy of $p+1-\alpha$ order. When $p=1$, i.e., using function values at both endpoints for interpolation, the method has the same numerical accuracy as the standard L1 scheme, which is $2-\alpha$ order. When $p=3$, i.e., using function values at both endpoints and first derivative values for interpolation, the method has a numerical accuracy of $4-\alpha$ order. When $p=5$, i.e., using function values at both endpoints, first derivative values, and second derivative values for interpolation, the method has a numerical accuracy of $6-\alpha$ order. 
 Due to the influence of floating-point errors and numerical issues during neural network optimization, the accuracy cannot reach the ideal state when $p=5$. Therefore, the commonly used case is $p=3$, i.e., using function values and first derivative values for interpolation. 
\end{remark}

\paragraph{The floating-point errors and numerical issues during neural network optimization}
Higher-order Hermite interpolation can generate high-degree polynomials, and the coefficients of these polynomials are dependent on the interval spacing of the interpolation region. The coefficients of the polynomial of degree $p$  generated by Hermite interpolation contain $\nicefrac{1}{\Delta t ^{p}}$, and these coefficients are involved in the calculations of neural networks. Therefore, when $p$ is large and $\Delta t$ is small, it is likely to cause floating-point errors and numerical issues during neural network optimization. 

Therefore, to ensure stable optimization of the neural network, we employ the double data type within the network. otherwise, it may lead to premature convergence, impacting the final accuracy. However, we do not need to be overly concerned about the additional time cost brought by double precision. Actually, from an experimental perspective, for a given equation, the model and data are not exceedingly large, and double precision does not significantly consume more time. 

Finally, these problems are also related to the computational process, and it might be possible to address this issue by simplifying expressions in advance to eliminate smaller values that arise during calculations. This is a consideration the authors are contemplating.

\section{Results}
In this section, we will demonstrate the performance of the HNS. We denote the HNS employing Hermite interpolation polynomial of degree $p$ as the $p$-th order  HNS. We denote the number of time nodes as $M_t=N+1$ and the time step as $\Delta t = \nicefrac{T}{N}$. The number of spatial nodes is denoted as $M_x$.
In our experiments, we evaluate the performance of the HNS by the relative L2 error $\nicefrac{\parallel \tilde{\boldsymbol{u}} - \boldsymbol{u}\parallel_{2}}{\parallel \boldsymbol{u}\parallel_{2}}$. Code is developed using PyTorch, the activate function $nn.GELU()$ \cite{hendrycks2016gaussian} and double data type are employed. 

We validate the high performance of HNS in six computational scenarios:(1)Fractional differential equation;(2)One-Dimensional Time-Fractional diffusion equation;(3)Two-Dimensional Time-Fractional advection-diffusion equation;(4)3-Dimensional Time-Fractional advection-diffusion equation;(5)10-Dimensional Time-Fractional advection-diffusion equation;(6)The Inverse Problem for 3-Dimensional Time-Fractional advection-diffusion equation.

\subsection{Fractional differential equation}
Consider the following fractional differential equation:
\begin{equation}
	{ }_{0}^{C} D_{t}^{\alpha} u(t) = u(t) + \frac{\Gamma(3)}{\Gamma(3-\alpha)} t^{2-\alpha} - t^{2} - 1,\ \  t \in [0,T],
\end{equation}
with the initial condition $u(0) = 1$. where $T$ is the time range. The exact solution for this equation is $u(t) = 1+t^2$. In this problem, we construct the trial solution:
 \begin{equation}
 	\tilde{u}(t) = t f(t;\theta) + 1,
 \end{equation}
 where $f(t;\theta)$ is a fully connected neural network with four hidden layers, each with 20 neurons. While at this point, the neural network only needs to learn a linear function, it is still possible to assess the performance of HNS. We study the performance of the HNS in handling this problem, and 1000 equally spaced test points used to calculate the prediction error. In the experiment, the time range $T$ is set to 2, and the number of time nodes $M_t$ is set to 3 and 6, respectively.
\begin{figure}[htbp]
  \centering
  \includegraphics[width=0.8\textwidth]{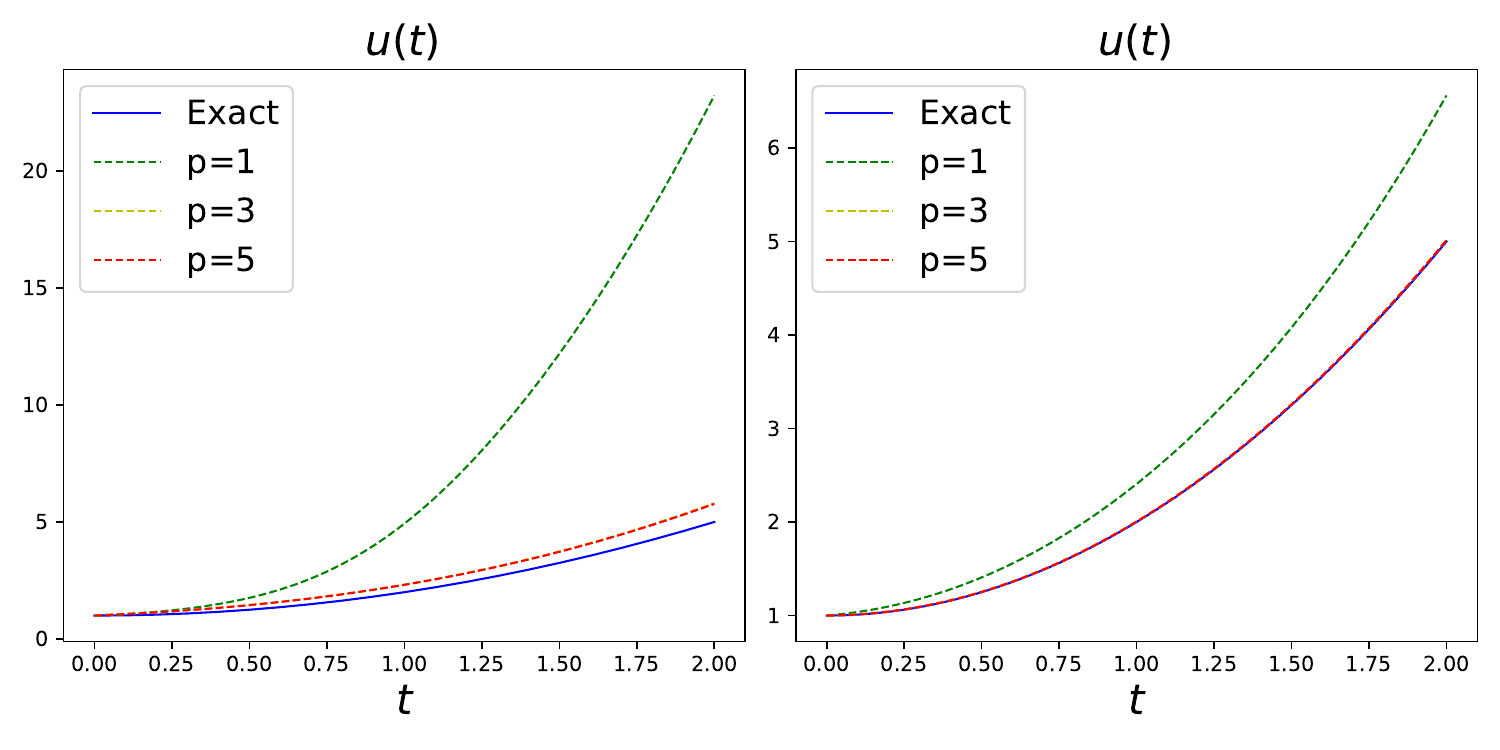}
  \caption{FDE: Comparison of the predicted solution and the exact solution when $\alpha=0.5$, with $p$ representing the $p$-th order HNS. left: $M_t = 3$ and right: $M_t=6$.}
  \label{fig:FDE}
\end{figure}
\begin{table}[htbp]
\caption{FDE: The relative L2 error for different time intervals.}
\label{FDE-table}
\centering
%\scriptsize
\begin{tabular}{llllllllll}
\toprule
& \multicolumn{3}{c}{$\alpha = 0.3$} & \multicolumn{3}{c}{$\alpha = 0.5$} & \multicolumn{3}{c}{$\alpha = 0.7$} \\
\cmidrule(r){2-4} \cmidrule(r){5-7} \cmidrule(r){8-10} 
$M_t$& $p=1$ & $p=3$ & $p=5$ & $p=1$ & $p=3$ & $p=5$ & $p=1$ & $p=3$ & $p=5$ \\
\midrule
6  & 1.83e-01 & 1.31e-03 & 1.37e-03 & 2.60e-01 & 2.82e-03 & 2.64e-03 & 3.82e-01 & 5.18e-03 & 5.14e-03\\
11  & 5.72e-02 & 6.92e-05 & 6.05e-05 & 8.65e-02 & 4.45e-05 & 6.27e-05 & 1.34e-01 & 8.54e-05 & 8.66e-05\\
21  & 1.88e-02 & 6.47e-05 & 6.90e-05 & 3.10e-02 & 1.24e-05 & 1.22e-05 & 5.24e-02 & 1.86e-05 & 1.86e-05\\
41  & 6.17e-03 & 8.94e-05 & 9.54e-05 & 1.12e-02 & 5.88e-06 & 1.15e-04 & 2.12e-02 & 7.15e-06 & 1.25e-04\\
81  & 2.04e-03 & 6.76e-05 & 8.72e-05 & 4.10e-03 & 4.70e-06 & 9.94e-05 & 8.62e-03 & 5.54e-06 & 9.65e-05\\
101  & 1.42e-03 & 6.03e-06 & 8.66e-05 & 2.95e-03 & 4.63e-06 & 9.42e-05 & 6.46e-03 & 5.01e-06 & 8.14e-05\\
\bottomrule
\end{tabular}
\end{table}

\reffig{fig:FDE} shows the comparison of the predicted and exact solutions when $\alpha=0.5$. It is important to note that the Hermite interpolation explicit approximation is equivalent to the traditional L1 scheme when $p=1$. The results indicate that the third-order or fifth-order HNS achieves improved predictive results when $M_t=3$ or $M_t=6$. For third-order HNS at $M_t=6$, the relative L2 error is reduced to as low as $2.82e-3$. Additionally, although a higher level of accuracy is theoretically expected for fifth-order HNS, this improvement is not significantly evident in practice. No significant differences in accuracy are observed between the third-order HNS and fifth-order HNS. Additionally, in terms of computational time, they are not as significant as initially anticipated. For specific details, please refer to \reftable{FDE_table_cost} in Appendix \ref{training_cost}.

Table \ref{FDE-table} presents the relative L2 errors of the model at $T=2$ for different time intervals. The table shows that the HNS achieves low errors for any value of $\alpha$. Additionally, both the third-order and fifth-order HNS show about 2 to 3 orders of magnitude higher accuracy than the HNS using linear Hermite interpolation. When the time interval is large, that is, when $M_t$ is small, the accuracy of the fifth-order HNS is similar to that of the third-order HNS. However, when the time interval is small, the fifth-order HNS actually reduces the accuracy. This is possibly due to floating-point errors and numerical issues during neural network optimization. Consequently, the third-order HNS provides the highest performance.

\begin{figure}[htbp]
  \centering
  \includegraphics[width=0.5\textwidth]{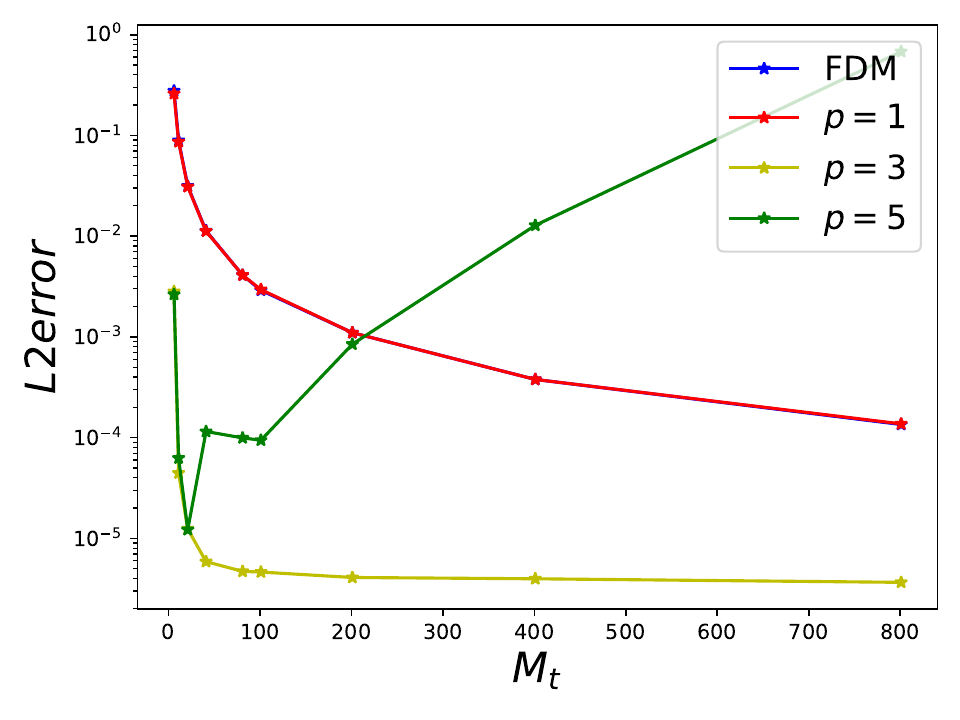}
  \caption{FDE: Comparison of the method proposed in this paper with the classical finite difference method based on the L1 schema in the time dimension when $\alpha=0.5$.}
  \label{fig:FDM_Comparison}
\end{figure}
Finally, we also compared HNS with finite difference numerical methods based on the L1 schema. \reffig{fig:FDM_Comparison} illustrates the accuracy comparison between standard FDM and different-order HNS for various time intervals. The results indicate that the accuracy of FDM is essentially equivalent to that of HNS with $p=1$, which is evident because when $p=1$, the fractional derivative approximation method used in HNS is equivalent to the L1 scheme. This also suggests that, given a numerical scheme, neural networks can learn the corresponding level of solution accuracy. Third-order HNS exhibits the highest precision, with accuracy of FDM dropping below $1e-3$ only after $M_t > 200$, while third-order HNS achieves errors below $1e-4$ by $M_t=11$. However, for fifth-order HNS, its accuracy is similar to third-order HNS when the time interval is relatively large. Still, as the time interval decreases, fifth-order HNS becomes unstable due to numerical issues, resulting in significant errors.

\subsection{Time-Fractional diffusion equation}
Next, let us consider the following time fractional diffusion equation:
\begin{equation}
	\left\{\begin{array}{l}
{ }_{0}^{C} D_{t}^{\alpha} u(x, t) = u_{xx}, \ \   x \in [0,L], t \in [0,T]\\
u(x, 0)=I(x), \ \ x \in [0,L] \\
u(0, t)=B(0,t),u(L, t)= B(L,t), \ \ t \in [0,T]
\end{array}\right.
\end{equation}
where $I(x) = x^{2}, B(0,t) = \frac{2 t^{\alpha}}{\Gamma(1+ \alpha)}$ and $ B(1,t) = 1+ \frac{2 t^{\alpha}}{\Gamma(1+ \alpha)}$. In this paper, we consider $L=1$ and $T=1$ as the spatial and temporal ranges, respectively.  The exact solution for this equation is 
$u(x,t) = x^{2}+ \frac{2 t^{\alpha}}{\Gamma(1+ \alpha)}$.
For this issue, we construct a trial solution
\begin{equation}
	\tilde{u}(x, t) = t f(x, t;\theta) + x^{2}
\end{equation}
where $f(x, t;\theta)$ represents a fully connected neural network with the number of layers and neurons consistent with previous experiments. 
\begin{figure}[htbp]
  \centering
  \includegraphics[width=1\textwidth]{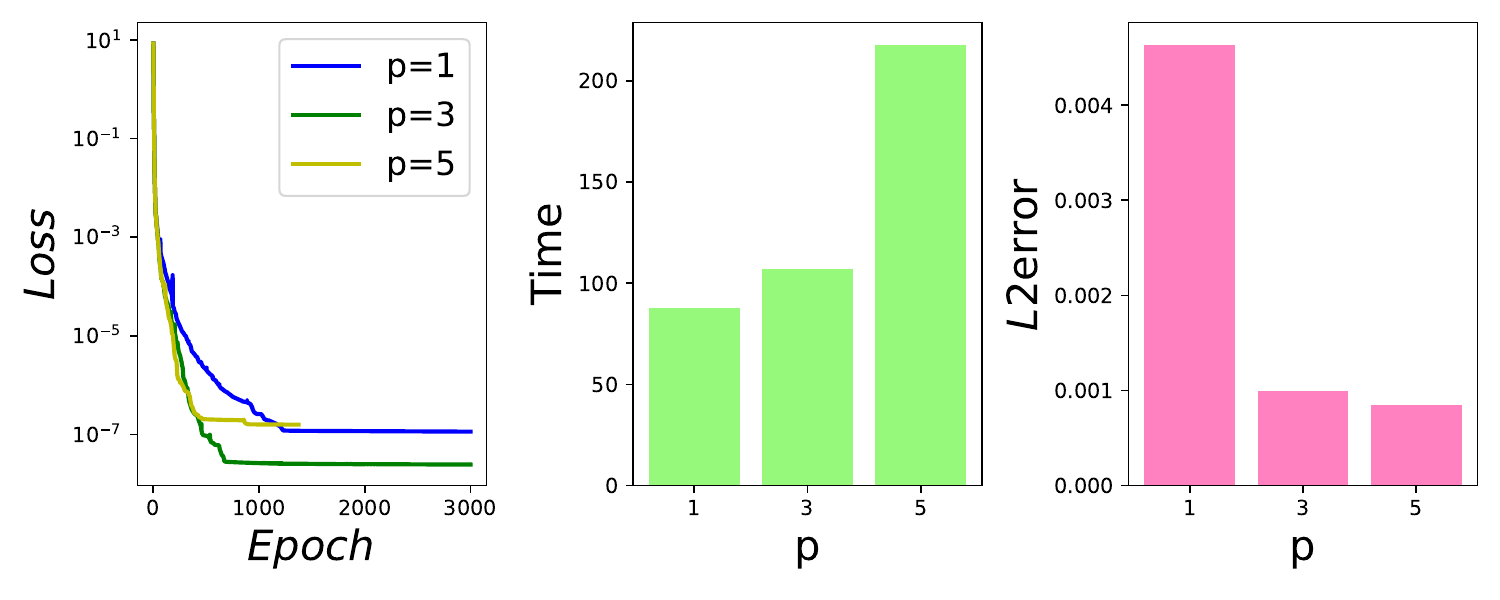}
  \caption{TFDE: The loss curve, computation time in seconds, and relative L2 error of the HNS with $\alpha=0.65$, $M_t=21$, and $M_x=11$.}
  \label{fig:TFDE_1}
\end{figure}
In this experiment, a uniform grid of $100 \times 100$ points is employed to test the model. \reffig{fig:TFDE_1} displays the performance of the HNS with $\alpha=0.65$, $M_t=21$, and $M_x=11$. The results indicate that higher-order HNS show faster convergence rates. As the convergence rate of the third-order HNS is faster, the time taken for convergence is quite close between the first-order and third-order HNS. However, the fifth-order HNS, after iterating over a thousand rounds, takes longer than the other models that iterate 3000 rounds. In terms of accuracy, both the third-order and fifth-order HNS demonstrate extremely high precision. Therefore, in terms of overall performance, the third-order HNS shows the highest level.

\begin{figure}[htbp]
  \centering
  \includegraphics[width=1\textwidth]{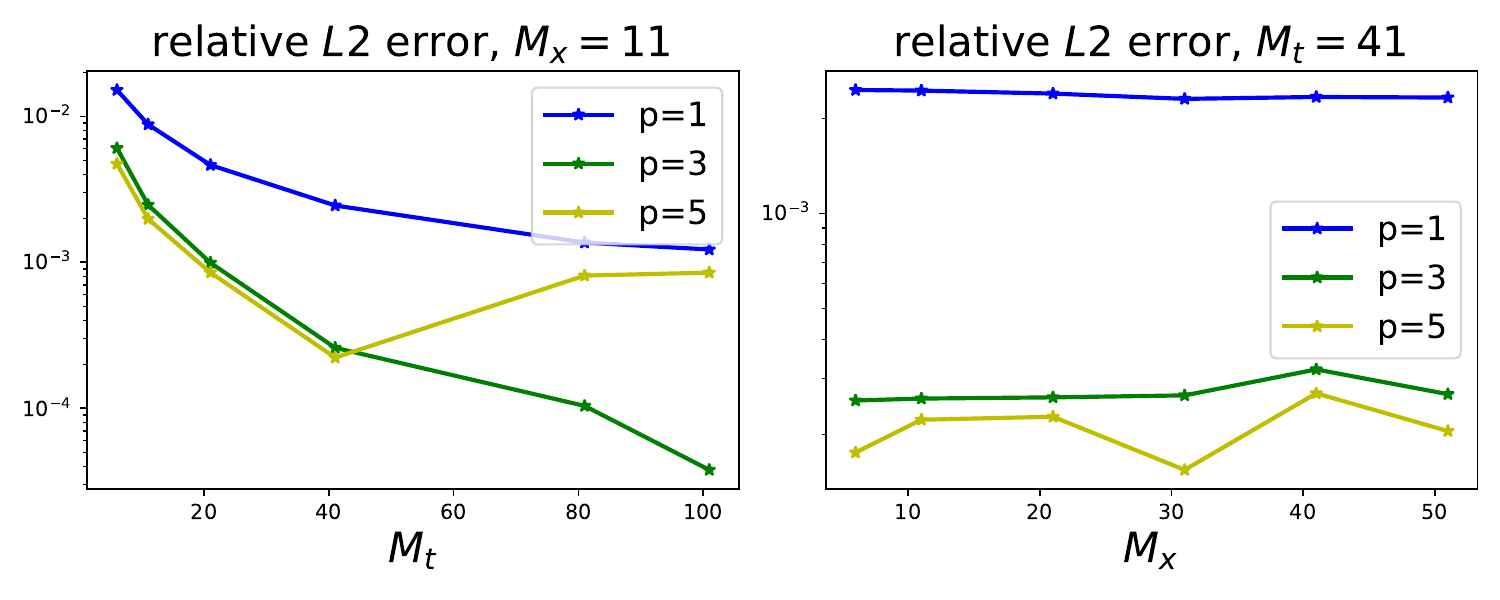}
  \caption{TFDE: The relative L2 error varies with changes in the number of time nodes $M_t$ or spatial nodes $M_x$, when $\alpha=0.65$.}
   \label{fig:TFDE_2}
\end{figure}

\reffig{fig:TFDE_2} illustrates the variations in relative L2 error with the number of temporal nodes $M_t$ or spatial nodes $M_x$ when $\alpha=0.65$. When $M_x$ is fixed at 11, the relative L2 errors of the first-order and third-order HNS gradually decrease as the time interval reduces. However, the accuracy of the fifth-order HNS shows a phenomenon of first decreasing and then increasing, which may be affected by floating-point errors and numerical issues during neural network optimization. In addition, when $M_t$ is fixed at 41, as $M_x$ increases, the accuracy of the first-order and third-order model does not show significant changes, while the relative L2 errors of the fifth-order HNS exists fluctuations. Both the third-order and fifth-order HNS have very high precision.

Table \ref{TFDE-table} displays the L2 error of the HNS model at different Caputo derivative orders and time intervals with $M_x=11$. It can be observed from the table that the HNS can solve the problem well for any $\alpha$. The third-order and fifth-order HNS 
 have higher accuracy.

\begin{table}[htbp]
\caption{TFDE: The relative L2 error for different time intervals.}
\label{TFDE-table}
\centering
%\scriptsize
\begin{tabular}{llllllllll}
\toprule
& \multicolumn{3}{c}{$\alpha = 0.45$} & \multicolumn{3}{c}{$\alpha = 0.65$} & \multicolumn{3}{c}{$\alpha = 0.85$} \\
\cmidrule(r){2-4} \cmidrule(r){5-7} \cmidrule(r){8-10} 
$M_t$& $p=1$ & $p=3$ & $p=5$ & $p=1$ & $p=3$ & $p=5$ & $p=1$ & $p=3$ & $p=5$ \\
\midrule
6  & 1.39e-02 & 1.73e-02 & 1.74e-02 & 1.52e-02 & 6.05e-03 & 4.73e-03 & 5.52e-03 & 1.57e-03 & 1.58e-03\\
11  & 9.79e-03 & 8.84e-03 & 8.40e-03 & 8.80e-03 & 2.47e-03 & 1.98e-03 & 4.01e-03 & 5.33e-04 & 5.32e-04\\
21  & 7.19e-03 & 3.81e-03 & 3.87e-03 & 4.64e-03 & 9.92e-04 & 8.51e-04 & 2.29e-03 & 1.73e-04 & 1.68e-04\\
41  & 2.80e-03 & 1.06e-03 & 1.22e-03 & 2.45e-03 & 2.60e-04 & 2.22e-04 & 1.33e-03 & 6.05e-05 & 9.24e-05\\
61  & 2.17e-03 & 2.76e-04 & 3.68e-04 & 1.71e-03 & 1.11e-04 & 1.70e-04 & 1.00e-03 & 5.53e-05 & 4.60e-04\\
81  & 1.73e-03 & 1.40e-04 & 3.44e-04 & 1.36e-03 & 1.04e-04 & 8.11e-04 & 8.86e-04 & 2.03e-05 & 4.47e-04\\
101  & 1.53e-03 & 1.66e-04 & 9.95e-04 & 1.22e-03 & 3.78e-05 & 8.50e-04 & 8.12e-04 & 4.25e-05 & 3.80e-04\\
\bottomrule
\end{tabular}
\end{table}

\subsection{Time-Fractional advection-diffusion equation}
In order to further illustrate the advantages of the model, we study the performance of the HNS in solving the two-dimensional fractional advection-diffusion equation. Consider the following fractional advection-diffusion equation:
\begin{equation}
	\left\{\begin{array}{l}
{ }_{0}^{C} D_{t}^{\alpha} u(\boldsymbol{x}, t) = \kappa \Delta  u(\boldsymbol{x}, t) + g(\boldsymbol{x},t), \ \   \boldsymbol{x} \in \Omega = [0,1]^{2}, t \in [0,1]\\
u(\boldsymbol{x}, 0)=0, \ \ \boldsymbol{x} \in \Omega\\
u(\boldsymbol{x}, t) = t^{2} e^{x+y}, \ \ \boldsymbol{x} \in \partial \Omega,t \in [0,1]
\end{array}\right.
\end{equation}
In this paper, we employ  $\kappa=1.0$ , and  $g(\boldsymbol{x}, t)=\left[\frac{2 t^{2-\alpha}}{\Gamma(3-\alpha)}-2 t^{2}\right] e^{x+y}$. The exact solution for this equation is 
	$u(\boldsymbol{x}, t)=t^{2} e^{x+y}$. For this case, we construct a trial solution:
 \begin{equation}
 	\tilde{u}(x,y, t) = t f(x, y, t;\theta),
 \end{equation}
In this experiment, we use a uniform grid of $100 \times 100 \times 100$ points to test the model. We first test the effectiveness of the HNS on this problem. With $\alpha=0.85, M_t = 11, M_x =11 \times 11$, we use the third-order HNS to solve the problem. The results show that the relative L2 error is $1.66e-04$, which demonstrates the high accuracy of the third-order HNS.

The performance of the HNS is evaluated with different values of $\alpha$, and the experimental results are presented in \reftable{TFADE-table}. It is observed that higher accuracy is achieved by the third-order and fifth-order HNS.
\begin{table}[htbp]
\caption{TFADE: The relative L2 error for different time intervals. $M_x =11 \times 11$}
\label{TFADE-table}
\centering
%\scriptsize
\begin{tabular}{llllllllll}
\toprule
& \multicolumn{3}{c}{$\alpha = 0.7$} & \multicolumn{3}{c}{$\alpha = 0.8$} & \multicolumn{3}{c}{$\alpha = 0.9$} \\
\cmidrule(r){2-4} \cmidrule(r){5-7} \cmidrule(r){8-10} 
$M_t$& $p=1$ & $p=3$ & $p=5$ & $p=1$ & $p=3$ & $p=5$ & $p=1$ & $p=3$ & $p=5$ \\
\midrule
6  & 6.05e-03 & 1.93e-04 & 2.80e-04 & 8.89e-03 & 3.13e-04 & 2.42e-04 & 1.13e-02 & 3.40e-04 & 3.64e-04\\
11  & 3.18e-03 & 1.73e-04 & 3.41e-04 & 4.02e-03 & 1.45e-04 & 2.11e-04 & 5.80e-03 & 1.74e-04 & 1.98e-04\\
21  & 1.70e-03 & 1.27e-04 & 1.31e-04 & 2.32e-03 & 1.98e-04 & 2.02e-04 & 3.31e-03 & 1.52e-04 & 1.49e-04\\
41  & 8.51e-04 & 1.06e-04 & 1.56e-04 & 1.32e-03 & 1.35e-04 & 1.93e-04 & 1.94e-03 & 1.68e-04 & 1.43e-04\\
81  & 4.41e-04 & 1.58e-04 & 2.70e-04 & 6.67e-04 & 1.62e-04 & 5.59e-04 & 1.06e-03 & 8.62e-05 & 8.97e-04\\
\bottomrule
\end{tabular}
\end{table}

\subsection{3D Time-Fractional advection-diffusion equation}
Now, let us address a higher-dimensional advection-diffusion equation. Firstly, we consider the following problem defined in a 3D cubic domain:
\begin{equation}
{ }_{0}^{C} D_{t}^{\alpha} u(\boldsymbol{x}, t)-\Delta u(\boldsymbol{x}, t) + (1,1,1) \cdot \nabla u(\boldsymbol{x}, t)-g(\boldsymbol{x}, t)=0, \ \ \ \boldsymbol{x} \in \Omega=(0,1)^{3}\in R^{3}
\end{equation}
The analytical solution for this equation is chosen as $u(\boldsymbol{x}, t) = t^2 + cos(x)+cos(y)+cos(z)$,with initial conditions $u(\boldsymbol{x}, 0)= cos(x)+cos(y)+cos(z)$. For this case, we construct a trial solution:
 \begin{equation}
 	\tilde{u}(x,y,z, t) = t f(x, y,z,t;\theta)+ cos(x)+cos(y)+cos(z),
 \end{equation}

In this experiment, we employ a uniform grid consisting of $51 \times 51 \times 51 \times 51$ points to evaluate the model. The time range is [0, 1], and $M_{x}=11 \times 11 \times 11$. The performance of the HNS is evaluated with $\alpha=0.5$, and the experimental results are presented in \reftable{FPDE-3D}. It is observed that higher accuracy is achieved by the third-order and fifth-order HNS. Moreover, the time expenses for the HNS models with $p=1$ and $p=3$ are comparable, while for $p=5$, convergence occurs earlier for larger time intervals, resulting in relatively shorter computation times. However, at $M_t=21$, the fifth-order HNS model takes 1186.06 seconds.

\begin{table}[htbp]
\caption{FPDE-3D: The relative L2 error and training cost for different time intervals.}
\label{FPDE-3D}
\centering
%\scriptsize
\begin{tabular}{llllllllll}
\toprule
& \multicolumn{6}{c}{$\alpha = 0.5$} \\
\cmidrule(r){2-7} 
$M_t$ &$p=1$ & cost(s)& $p=3$& cost(s)& $p=5$& cost(s) \\
\midrule
3 & 4.41e-03 & 149.21 & 1.72e-03&153.96 & 1.60e-03&168.77 \\
6   & 4.57e-04 & 232.78 & 5.86e-05&239.83 & 5.93e-05&277.49 \\
11  & 2.23e-04 &370.09&3.62e-05&395.34 & 2.32e-05&514.28 \\
21  & 9.66e-05 &860.74&2.94e-05&881.57 & 3.09e-05&1186.06  \\
\bottomrule
\end{tabular}
\end{table}

In this experiment, we construct a trial solution that is suitable. So the neural network only needs to learn a linear function, which simplifies the problem. To thoroughly verify the performance of HNS, we change the analytical solution for this problem to $u(\boldsymbol{x}, t) = t^2(cos(x_1)+cos(x_2)+cos(x_3))$,with initial conditions $u(\boldsymbol{x}, 0)= 0$. Now, we construct a trial solution:
 \begin{equation}
 	\tilde{u}(x_1,x_2,x_3,t) = t f(x_1,x_2,x_3,t;\theta),
 \end{equation}

Additionally, it is worth noting that using equidistant grid points for training models in high-dimensional problems is not suitable. This is because as the dimensionality increases, the number of equidistant points grows exponentially. Therefore, to demonstrate the ability of HNS to handle high-dimensional problems, we utilize uniform sampling to generate spatial coordinates both inside the domain and on the boundaries. This approach helps prevent issues associated with the explosion of dimensions. In this paper, we employ the LHS method to sample training spatial coordinate points.

We uniformly sample 5000000 spatiotemporal coordinates to assess the performance of the model. For the training data, we uniformly sample 1000 coordinates in space, i.e., $M_x=1000$, with 100 coordinates uniformly sampled along each boundary. At $M_t=6$, the relative L2 error of the first-order HNS model is $2.47e-03$. But the relative L2 error of the third-order HNS is $3.66e-04$, which demonstrates the high accuracy of the third-order HNS.

\subsection{10D Time-Fractional advection-diffusion equation}
Next, we extend the equation from 3D to 10D. We choose the analytical solution for the problem as follows:
\begin{equation}
\label{eq:e38}
	u(\boldsymbol{x}, t) = t^2 \sum_i^N cos(x_i)
\end{equation}
where $N=10$. The initial condition remains $u(\boldsymbol{x}, 0)= 0$, so the constructed trial solution is:
 \begin{equation}
 \label{eq:e39}
 	\tilde{u}(\boldsymbol{x},t) = t f(\boldsymbol{x},t;\theta),
 \end{equation}
Firstly, we conduct testing on a 10-dimensional advection equation, which takes the following form:
 \begin{equation}
{ }_{0}^{C} D_{t}^{\alpha} u(\boldsymbol{x}, t)+ (1,1,1) \cdot \nabla u(\boldsymbol{x}, t)-g(\boldsymbol{x}, t)=0, \ \ \ \boldsymbol{x} \in \Omega=(0,1)^{10}\in R^{10}
\end{equation}
 
 We continue to employ 5000000 spatiotemporal coordinates to assess the performance of the model. For the training data, we uniformly sample 5000 coordinates in space, i.e., $M_x=5000$, with 100 coordinates uniformly sampled along each boundary. Due to the complexity of the problem, we set the number of training epochs to 3000, and the specific experimental results are shown in Table \ref{FPDE-10D_adv}. The experimental results indicate that both $p=1$ and $p=3$ HNS models can effectively solve this problem. Meanwhile, the third-order HNS exhibits higher accuracy, and this accuracy does not decrease as the time interval decreases, suggesting that the accuracy has likely reached the limit for this problem. Additionally, the third-order HNS does not show a significant increase in computational time.
 \begin{table}[htbp]
\caption{10D advection equation: The relative L2 error and training cost for different time intervals.}
\label{FPDE-10D_adv}
\centering
%\scriptsize
\begin{tabular}{llllllllll}
\toprule
& \multicolumn{4}{c}{$\alpha = 0.5$} \\
\cmidrule(r){2-5} 
$M_t$ &$p=1$ & cost(s)& $p=3$& cost(s) \\
\midrule
6   & 5.78e-03 & 356.35 & 2.02e-04 & 413.42  \\
11  & 2.26e-03 & 538.09 & 2.34e-04 & 773.69  \\
21  & 9.19e-04 & 2382.85& 2.16e-04 & 2434.15   \\
\bottomrule
\end{tabular}
\end{table}

\begin{figure}[htbp]
  \centering
  \includegraphics[width=1\textwidth]{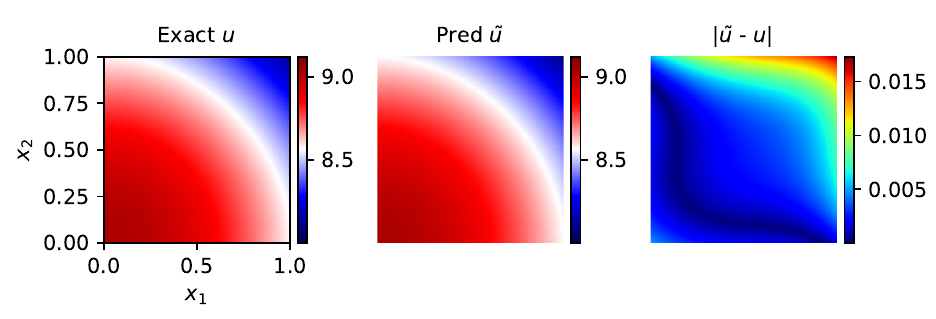}
  \caption{10D Time-Fractional advection-diffusion equation when $\alpha=0.5$.}
   \label{fig:error10D}
\end{figure}

Next, we will upgrade the experiment from an advection equation to an advection-diffusion equation, which takes the following form:
\begin{equation}
{ }_{0}^{C} D_{t}^{\alpha} u(\boldsymbol{x}, t)-\Delta u(\boldsymbol{x}, t) + (1,1,1) \cdot \nabla u(\boldsymbol{x}, t)-g(\boldsymbol{x}, t)=0, \ \ \ \boldsymbol{x} \in \Omega=(0,1)^{10}\in R^{10}
\end{equation}
We continue to employ \refeq{eq:e38} as the analytical solution for the problem and \refeq{eq:e39} as the trial solution for HNS. The settings for testing and training data remain the same as before. Since the equation includes spatial second-order derivatives, the model will take more time. However, the model still maintains high accuracy, and its precision does not decrease due to the complexity of the equation. At $M_t=6$, the relative L2 error of the first-order HNS model is $1.37e-03$, with a cost of 1074.03. However, the relative L2 error of the third-order HNS is $8.26e-04$, with a cost of 1077.11. \reffig{fig:error10D} illustrates the results obtained by the third-order HNS, corresponding to $t=1, x_3 = x_4= ... = x_{10} = 0.5$. From the figure, it is evident that the HNS results exhibit very small errors when compared to the analytical solution.

\subsection{The Inverse Problem}
We have already thoroughly demonstrated the excellent performance of HNS on forward problems. In this section, we shift our focus to solving inverse problems. Let us reconsider the 3D advection-diffusion equation:
\begin{equation}
{ }_{0}^{C} D_{t}^{\alpha} u(\boldsymbol{x}, t)-\beta \Delta u(\boldsymbol{x}, t) + \gamma (1,1,1) \cdot \nabla u(\boldsymbol{x}, t)-g(\boldsymbol{x}, t)=0, \ \ \ \boldsymbol{x} \in \Omega=(0,1)^{3}\in R^{3}
\end{equation}
along with its analytical solution $u(\boldsymbol{x}, t) = t^2(cos(x_1)+cos(x_2)+cos(x_3))$ and initial conditions $u(\boldsymbol{x}, 0)= 0$. Additionally, $g(\boldsymbol{x}, t)$ is known. In the inverse problem, the goal is to infer $(\alpha, \beta ,\gamma)$ from a given measurement of $u(\boldsymbol{x}, t)$.

First, let us clarify our experimental data. In this problem, we uniformly sample 5000000 spatiotemporal coordinates to assess the performance of the model. For the training data, we uniformly sample 1000 coordinates in space, i.e., $M_x=1000$, with 50 coordinates uniformly sampled along each boundary. In the temporal dimension, we set $M_t=6$. Additionally, in our model, we treat $(\alpha, \beta ,\gamma)$ as optimizable parameters that are optimized together with the parameters of neural network. To solve this inverse problem, we need to include an additional term in the loss function:
 \begin{gather}
 	\mathcal{L}(\theta,\alpha, \beta ,\gamma)= \frac{1}{N_{b}} \sum_{j=1}^{N_{b}}\left[\tilde{u}(\boldsymbol{x}_{b}^{j}, t_{b}^{j})-B(\boldsymbol{x}_{b}^{j}, t_{b}^{j})\right]^{2}
	 +\frac{1}{N_{r}} \sum_{j=1}^{N_{r}}\left[r(\boldsymbol{x}_{r}^{j}, t_{r}^{j})\right]^{2}
	 +\frac{1}{N_{r}} \sum_{j=1}^{N_{r}}\left[\tilde{u}(\boldsymbol{x}_{r}^{j}, t_{r}^{j})-u(\boldsymbol{x}_{r}^{j}, t_{r}^{j})\right]^{2}
	  \end{gather}
Firstly, we assume that there is only one unknown parameter $\alpha$ in the equation. We use HNS with $p=1$ and $p=3$ to solve this problem, respectively. We set the initial value of $\alpha$ to $0.2$, while its true value is $0.5$. Table \ref{Inverse_3D_1} indicates that the third-order HNS exhibits a significant advantage both in parameter inference and solution accuracy compared to $p=1$. Moreover, their computational time difference is not substantial. In this problem, the error in $\alpha$ is only $0.00486$.

\begin{table}[htbp]
\caption{The Inverse Problem for 3D advection-diffusion equation to infer $\alpha$.}
\label{Inverse_3D_1}
\centering
%\scriptsize
\begin{tabular}{llllllllll}
\toprule
& \multicolumn{4}{c}{$\alpha = 0.5$} \\
\cmidrule(r){2-5} 
$p$ &$\tilde{\alpha}$ & $|\tilde{\alpha}-\alpha|$ & $\nicefrac{\parallel \tilde{\boldsymbol{u}} - \boldsymbol{u}\parallel_{2}}{\parallel \boldsymbol{u}\parallel_{2}}$ & cost(s) \\
\midrule
1  & 0.54235 & 0.04235 & 1.03e-03 & 198.70  \\
3  & 0.49514 & 0.00486 & 2.47e-04 & 214.79  \\
\bottomrule
\end{tabular}
\end{table}
Next, we expand the parameter inference scope and use the HNS model to infer $(\alpha, \beta ,\gamma)$,setting their initial values to $(0.2, 0.2 ,0.2)$. From Table \ref{Inverse_3D_2}, it is evident that HNS has performed admirably in solving the inverse problem, with very small errors for each unknown parameter. Furthermore, the third-order HNS demonstrates superior accuracy. Surprisingly, the error in $\alpha$ has even decreased in this case.

\begin{table}[htbp]
\caption{The Inverse Problem for 3D advection-diffusion equation to infer $(\alpha, \beta ,\gamma)$.}
\label{Inverse_3D_2}
\centering
%\scriptsize
\begin{tabular}{llllllllll}
\toprule
& \multicolumn{8}{c}{$\alpha = 0.5, \beta = 1.0,\gamma = 1.0$} \\
\cmidrule(r){2-9} 
$p$ &$\tilde{\alpha}$ & $|\tilde{\alpha}-\alpha|$ &$\tilde{\beta}$ & $|\tilde{\beta}-\beta|$ &$\tilde{\gamma}$ & $|\tilde{\gamma}-\gamma|$ & $\nicefrac{\parallel \tilde{\boldsymbol{u}} - \boldsymbol{u}\parallel_{2}}{\parallel \boldsymbol{u}\parallel_{2}}$ & cost(s) \\
\midrule
1  & 0.46707 & 0.03293 &1.14661& 0.14661 &1.10838& 0.10838&1.46e-03 & 201.27 \\
3  & 0.49714 & 0.00286 & 0.96709 & 0.03291 &0.95065& 0.04935 & 2.75e-04 & 209.07  \\
\bottomrule
\end{tabular}
\end{table}

\section{Conclusion and Discussion}
In this paper, we extend the explicit approximation of Caputo fractional derivatives based on Hermite interpolation technique and analyze the approximation error. By utilizing the infinitely differentiable properties of deep neural networks, we naturally combine the Hermite interpolation explicit approximation of Caputo fractional derivatives with deep neural networks to propose an innovative model, called HNS. This model can solve time-fractional partial differential equations with high accuracy. Experimental results show that the accuracy of the third-order HNS is significantly higher than that of the neural network method based on L1 scheme for both forward and inverse problems, as well as in high-dimensional scenarios.

The HNS fully utilizes the advantages of neural networks and breaks through the limitation of explicit finite difference methods that can only be interpolated based on function values. This results in deeper integration models between deep neural networks and numerical methods, and opens up new opportunities for integrating deep neural networks with scientific computing technology. The limitation of this work is we focus exclusively on time-fractional partial differential equations. Exploring spatial fractional derivatives and other important fractional partial differential equations represents an exciting direction for future research.

\section*{Declaration of competing interest}
The authors declare that they have no known competing financial interests or personal relationships that could have appeared to influence the work reported in this paper.

\section*{Data availability}
The data and code can be found at \url{https://github.com/hsbhc/HNS}.

\section*{Acknowledgements}
This work is supported by the National Key Research and
Development Program of China (No. 2021YFA1003004).

\section*{Appendix.}
\appendix
\section{ Approximation Error of Hermite Interpolation Explicit Approximation}
\noindent {\bf Theorem 2.} Suppose $u \in C^{p+1}(t_{0},t_{n})$,then the approximation error is given by:
\begin{equation}
\begin{aligned}
	&\mid R(u(t_{n})) \mid \\
	&\leq \frac{1}{(p+1) !\Gamma(1-\alpha)} \left[ \frac{1}{2^{p+1}} + \alpha \frac{\Gamma(\frac{3+p}{2}) \Gamma(\frac{1-2\alpha+p}{2})}{\Gamma(2-\alpha+p)} \right]\underset{t_{0}\leq \xi \leq t_{n}}{\operatorname{max}} \left|u^{(p+1)}(\xi)\right| \Delta t^{p+1-\alpha}
\end{aligned}
\end{equation}
\begin{proof}
\label{Proof}
From the definition of $R(u(t_n))$, we know that:
\begin{equation}
\begin{aligned}
	R(u(t_{n})) &= \frac{1}{\Gamma(1-\alpha)} \sum_{k=1}^{n}\int_{t_{k-1}}^{t_{k}} \frac{u^{\prime}(t)}{(t_{n}-t)^{\alpha}} dt - \frac{1}{\Gamma(1-\alpha)} \sum_{k=1}^{n}\int_{t_{k-1}}^{t_{k}} \frac{H^{\prime}_{p}(t)}{(t_{n}-t)^{\alpha}} dt\\
	& = \frac{1}{\Gamma(1-\alpha)} \sum_{k=1}^{n}\int_{t_{k-1}}^{t_{k}} \frac{\left[u(t) - H_{p}(t)\right]^{\prime}}{(t_{n}-t)^{\alpha}} dt
\end{aligned}
\end{equation}
Applying the integration by parts formula and noting the error function of Hermite Interpolation, we obtain:
\begin{equation}
\begin{aligned}
	R(u(t_{n})) &= - \frac{1}{\Gamma(1-\alpha)} \sum_{k=1}^{n}\int_{t_{k-1}}^{t_{k}} \left[u(t) - H_{p}(t)\right]d{\frac{1}{(t_{n}-t)^{\alpha}}}\\
	& = - \frac{1}{\Gamma(1-\alpha)} \sum_{k=1}^{n}\int_{t_{k-1}}^{t_{k}} \frac{u^{(p+1)}(\xi)}{(p+1) !} \prod_{i=k-1}^{k}\left(t-t_{i}\right)^{\frac{(p+1)}{2}} \alpha (t_{n}-t)^{-\alpha-1} dt
	\end{aligned}
\end{equation}
Thus,
\begin{equation}
\label{eq48}
	\mid R(u(t_{n})) \mid \leq \frac{1}{(p+1) !\Gamma(1-\alpha)} \underset{t_{0}\leq \xi \leq t_{n}}{\operatorname{max}} \left|u^{(p+1)}(\xi)\right| \sum_{k=1}^{n}\int_{t_{k-1}}^{t_{k}} \prod_{i=k-1}^{k}
	 (\mid t-t_{i}\mid)^{\frac{(p+1)}{2}} \alpha (t_{n}-t)^{-\alpha-1} dt
\end{equation}
Two formulas can be obtained by calculation as follows:
\begin{equation}
\begin{aligned}
\label{eq49}
	& \sum_{k=1}^{n-1}\int_{t_{k-1}}^{t_{k}}
	 (t-t_{k-1})^{\frac{(p+1)}{2}} (t_{k}-t)^{\frac{(p+1)}{2}} \alpha (t_{n}-t)^{-\alpha-1} dt \\
	 &  \leq \frac{\Delta t ^{p+1}}{2^{p+1}} \sum_{k=1}^{n-1}\int_{t_{k-1}}^{t_{k}} \alpha (t_{n}-t)^{-\alpha-1} dt\\
	 & = \frac{\Delta t ^{p+1}}{2^{p+1}} \int_{t_{0}}^{t_{n-1}}\alpha (t_{n}-t)^{-\alpha-1} dt\\
	 & = \frac{\Delta t ^{p+1}}{2^{p+1}} (\Delta t^{-\alpha} - t_{n}^{-\alpha}) \leq \frac{1}{2^{p+1}} \Delta t ^{p+1-\alpha}
\end{aligned}
\end{equation}
and:
\begin{equation}
\label{eq50}
\begin{aligned}
	& \int_{t_{n-1}}^{t_{n}}
	 (t-t_{n-1})^{\frac{(p+1)}{2}} (t_{n}-t)^{\frac{(p+1)}{2}} \alpha (t_{n}-t)^{-\alpha-1} dt \\
	 &  \alpha \int_{t_{n-1}}^{t_{n}}
	 (t-t_{n-1})^{\frac{(p+1)}{2}} (t_{n}-t)^{\frac{(p+1)}{2} - \alpha -1}dt \\
	 	 & = \alpha \int_{0}^{\Delta t}
	 (\Delta t- \xi)^{\frac{(p+1)}{2}} \xi ^{\frac{(p+1)}{2} - \alpha -1}d\xi \\
	 & = \alpha \frac{\Gamma(\frac{3+p}{2})\Gamma(\frac{1-2\alpha+p}{2})}{\Gamma(2-\alpha+p)} \Delta t^{p+1-\alpha}
\end{aligned}
\end{equation}
Substituting \refeq{eq49} and \refeq{eq50} into \refeq{eq48}, we obtain:
\begin{equation}
\begin{aligned}
	&\mid R(u(t_{n})) \mid \\
	&\leq \frac{1}{(p+1) !\Gamma(1-\alpha)} \left[ \frac{1}{2^{p+1}} + \alpha \frac{\Gamma(\frac{3+p}{2}) \Gamma(\frac{1-2\alpha+p}{2})}{\Gamma(2-\alpha+p)} \right]\underset{t_{0}\leq \xi \leq t_{n}}{\operatorname{max}} \left|u^{(p+1)}(\xi)\right| \Delta t^{p+1-\alpha}
\end{aligned}
\end{equation}
\end{proof}

\section{Additional results}
In this section, we present additional results that help clarify details of our method.
\subsection{Fractional differential equation}

\begin{figure}[htbp]
  \centering
  \includegraphics[width=0.4\textwidth]{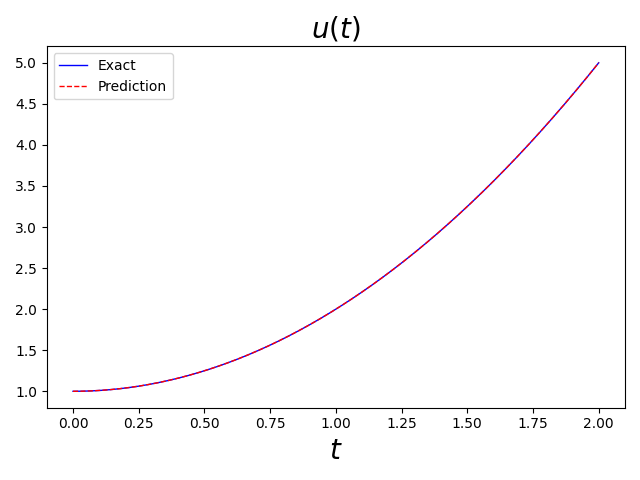}
  \includegraphics[width=0.4\textwidth]{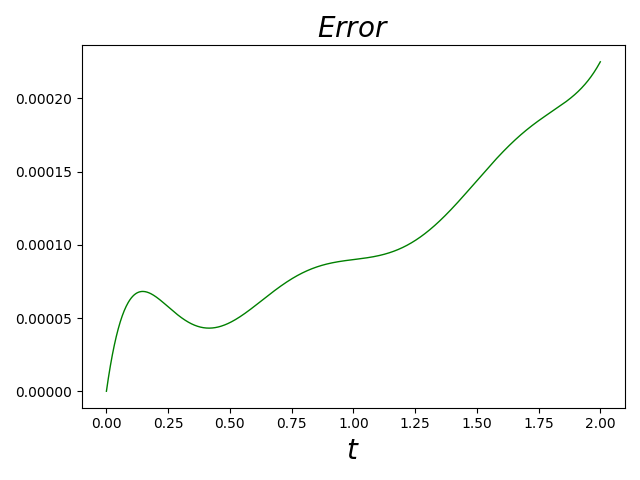}
  \caption{FDE: Comparison between predicted and exact solutions, the absolute error at each time step with the third-order HNS, using parameters $\alpha=0.5$, $M_t=11$, and $p=3$.}
\label{fig:FDE_add_1}
\end{figure}
\begin{figure}[htbp]
  \centering
  \includegraphics[width=0.5\textwidth]{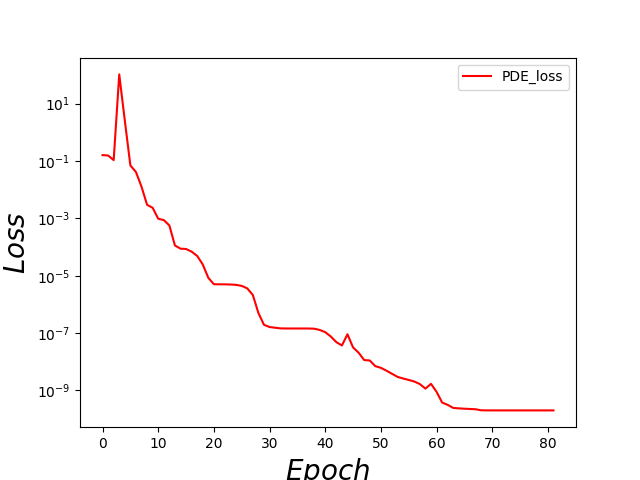}
  \caption{FDE: The iterative decrease in training loss with the third-order HNS, using parameters $\alpha=0.5$, $M_t=11$, and $p=3$.}
\label{fig:FDE_add_3}
\end{figure}

\reffig{fig:FDE_add_1} and \reffig{fig:FDE_add_3} demonstrate the predictive results and convergence process of the third-order HNS. During the training process, a large time interval of $M_t = 11$ is utilized, yet the relative L2 error of the third-order HNS is as low as $4.45e-05$. And the absolute error increases as $t$ increases. This effectively illustrates the high accuracy of the third-order HNS. Upon examining the loss curve, it is evident that the convergence rate of the third-order HNS is remarkably fast. 

\paragraph{training cost}
\label{training_cost}
Due to the need to compute second-order derivatives, the computational cost for $p=5$ is considerably high. Theoretically, the time cost follows the order: $p=1 < p=3 < p=5$. However, $p=3$ and $p=5$ converge faster because they utilize function values, derivatives, or higher-order derivatives for interpolation. This essentially compensates for the disadvantage in terms of computational time, possibly resulting in $p=3$ and $p=5$ having lower time costs compared to $p=1$. See \reftable{FDE_table_cost}
\begin{table}[htbp]
\caption{FDE: The relative L2 error and training cost. Training 3000 epochs.}
\label{FDE_table_cost}
\centering
%\scriptsize
\begin{tabular}{lllllllllllllll}
\toprule
& \multicolumn{6}{c}{$\alpha = 0.5$} \\
\cmidrule(r){2-7} 
$M_t$ &$p=1$ & cost(s)& $p=3$& cost(s)& $p=5$& cost(s) \\
\midrule
6   & 2.60e-01 &\textcolor{blue}{0.20} & 2.82e-03&\textcolor{blue}{0.33} & 2.64e-03&\textcolor{blue}{0.91} \\
11  & 8.65e-02 &10.69&4.45e-05&\textcolor{blue}{0.61} & 6.27e-05&104.84 \\
21  & 3.10e-02 &15.38&1.24e-05&27.95 & 1.22e-05&\textcolor{blue}{6.99}  \\
41  & 1.12e-02 &22.92&5.88e-06& 48.31 & 1.15e-04&\textcolor{blue}{7.30} \\
81  & 4.10e-03 &36.62 &4.70e-06&\textcolor{blue}{2.29} & 9.94e-05&\textcolor{blue}{12.07} \\
101 & 2.95e-03 &45.92&4.63e-06&\textcolor{blue}{4.80} & 9.42e-05&\textcolor{blue}{13.56}\\
\bottomrule
\end{tabular}
\end{table}

\subsection{Time-Fractional diffusion equation}
\begin{figure}[htbp]
  \centering
  \includegraphics[width=0.4\textwidth]{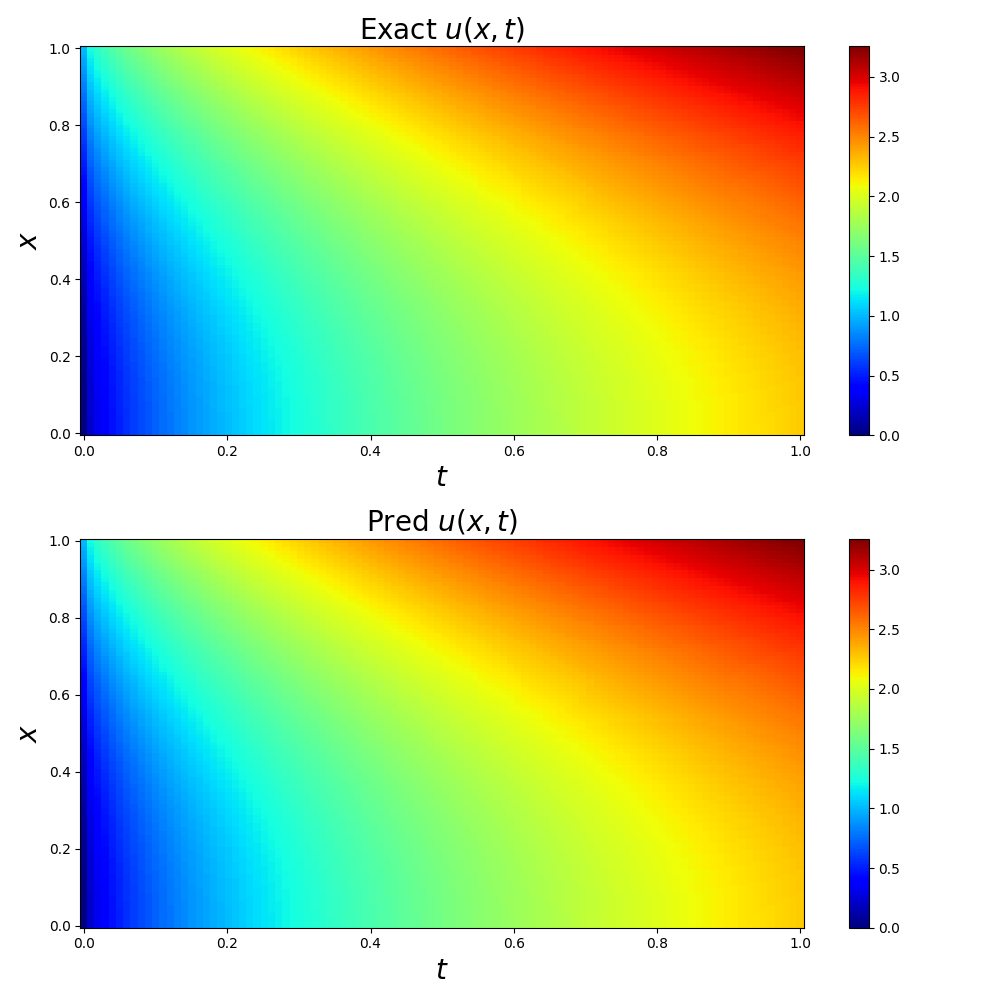}
  \caption{TFDE: Comparison between predicted and exact solutions with the third-order HNS, using parameters $\alpha=0.5$, $M_t=51, M_x = 11$.}
\label{fig:TFDE_add_1}
\end{figure}

\begin{figure}[htbp]
  \centering
  \includegraphics[width=0.4\textwidth]{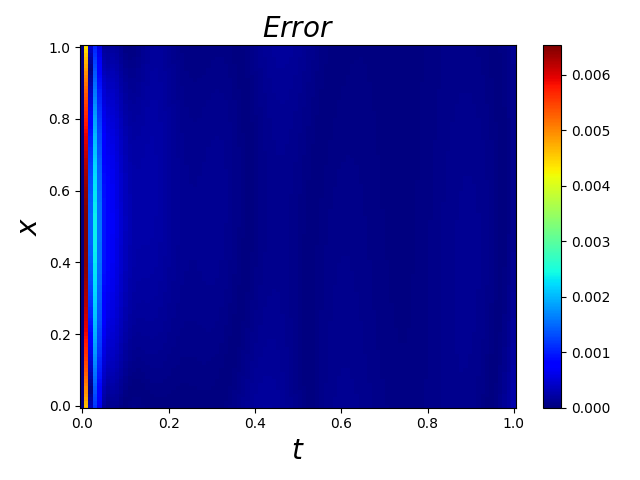}
  \includegraphics[width=0.4\textwidth]{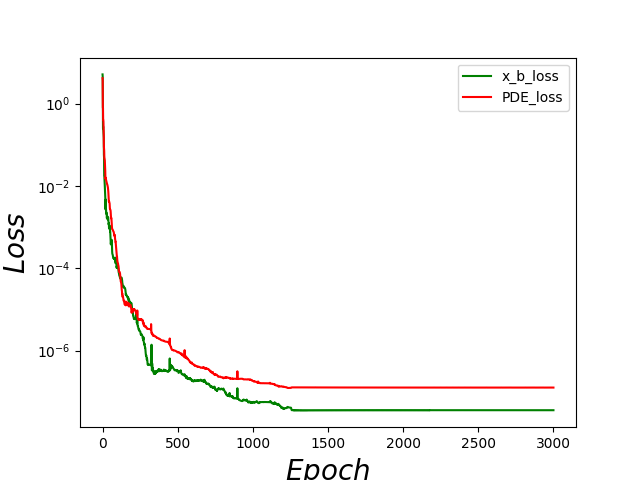}
  \caption{TFDE:The absolute error, and the iterative decrease in training loss with the third-order HNS, using parameters $\alpha=0.5$, $M_t=51, M_x = 11$.}
\label{fig:TFDE_add_2}
\end{figure}

\reffig{fig:TFDE_add_1} and \reffig{fig:TFDE_add_2} illustrate the predictive results and convergence process of the third-order HNS, which achieves a remarkably low relative L2 error of $3.39e-04$ for time-fractional PDEs. Unlike previous experiments, we observe that the absolute error does not increase with time for this problem. As shown in the figure, the error is large near $t=0$ but relatively small for later time steps. This phenomenon may be related to the optimization process of the third-order HNS, and we will further analyze it using Neural Tangent Kernel(NTK) in future work.

\subsection{Time-Fractional advection-diffusion equation}

\begin{figure}[htbp]
  \centering
  \includegraphics[width=0.5\textwidth]{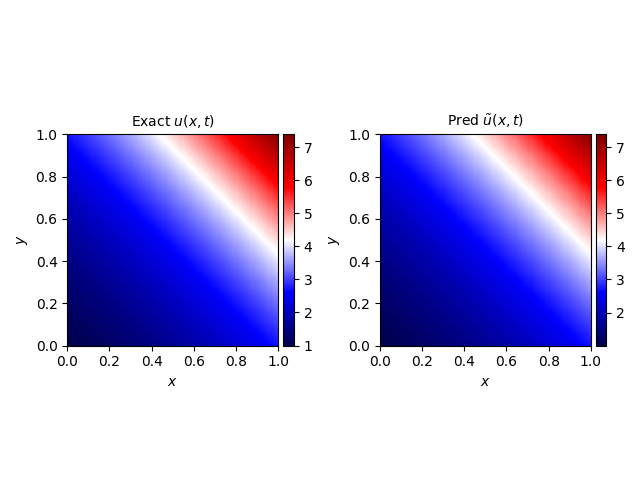}
    \caption{TFADE: Comparison between predicted and exact solutions of $t=1$ with the third-order HNS, using parameters $\alpha=0.85$, $M_t=11, M_x = 11\times 11$.}
\label{fig:TFADE_add_2}
\end{figure}

\begin{figure}[htbp]
  \centering
  \includegraphics[width=0.5\textwidth]{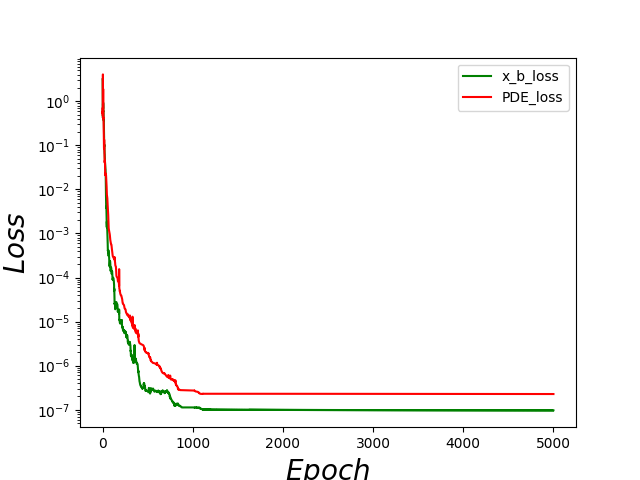}
    \caption{TFADE: The iterative decrease in training loss with the third-order HNS, using parameters $\alpha=0.85$, $M_t=11, M_x = 11\times 11$.}
\label{fig:TFADE_add_1}
\end{figure}

\reffig{fig:TFADE_add_2} and \reffig{fig:TFADE_add_1} illustrates the predictive results of the third-order HNS, which achieves a remarkably low relative L2 error of $1.66e-04$ for this problem. 

\bibliographystyle{model1-num-names}
\bibliography{Ref}
\end{document}